%% file: neurips_2024.tex
\pdfoutput=1
\documentclass{article}

\PassOptionsToPackage{numbers, compress, sort}{natbib}
\usepackage[final]{neurips_2024} 

\usepackage[utf8]{inputenc} 
\usepackage[T1]{fontenc}    
\usepackage{hyperref}       
\usepackage{url}            
\usepackage{booktabs}       
\usepackage{amsfonts}       
\usepackage{nicefrac}       
\usepackage{microtype}      
\usepackage{xcolor}         

\usepackage{float}
\usepackage{stfloats}
\usepackage{graphicx}
\usepackage{algorithm}
\usepackage[noend]{algpseudocode}
\usepackage{amsthm,amsmath, amssymb}
\usepackage{booktabs, ctable}
\usepackage{multirow}

\title{Pretraining with Random Noise for Fast and Robust Learning without Weight Transport}

\author{
  Jeonghwan Cheon$^{1}$ \quad Sang Wan Lee$^{1,2,3}$ \quad Se-Bum Paik$^{1}$\\
  $^1$Department of Brain and Cognitive Sciences, \\
  $^2$Graduate School of Data Science, $^3$Kim Jaechul Graduate School of AI\\
  Korea Advanced Institute of Science and Technology, Daejeon, Republic of Korea \\
  \texttt{\{jeonghwan518, sangwan, sbpaik\}@kaist.ac.kr}\\
}

\begin{document}

\maketitle
\vspace{-0.2cm}
\begin{abstract}
  The brain prepares for learning even before interacting with the environment, by refining and optimizing its structures through spontaneous neural activity that resembles random noise. However, the mechanism of such a process has yet to be understood, and it is unclear whether this process can benefit the algorithm of machine learning. Here, we study this issue using a neural network with a feedback alignment algorithm, demonstrating that pretraining neural networks with random noise increases the learning efficiency as well as generalization abilities without weight transport. First, we found that random noise training modifies forward weights to match backward synaptic feedback, which is necessary for teaching errors by feedback alignment. As a result, a network with pre-aligned weights learns notably faster and reaches higher accuracy than a network without random noise training, even comparable to the backpropagation algorithm. We also found that the effective dimensionality of weights decreases in a network pretrained with random noise. This pre-regularization allows the network to learn simple solutions of a low rank, reducing the generalization error during subsequent training. This also enables the network to robustly generalize a novel, out-of-distribution dataset. Lastly, we confirmed that random noise pretraining reduces the amount of meta-loss, enhancing the network ability to adapt to various tasks. Overall, our results suggest that random noise training with feedback alignment offers a straightforward yet effective method of pretraining that facilitates quick and reliable learning without weight transport.
\end{abstract}
\vspace{-0.2cm}

\input{section/introduction.tex}
\input{section/preliminaries.tex}
\input{section/methods.tex}
\input{section/results.tex}
\input{section/discussion.tex}
\input{section/limitations_and_broader_impacts.tex}

\section{Code availability}
Python 3.11 (Python software foundation) with PyTorch 2.1 was used to perform the simulation. The code used in this work is available at \url{https://github.com/cogilab/Random}.

\newpage

\section*{Acknowledgements}

This work was supported by the National Research Foundation of Korea (NRF-2022R1A2C3008991 to S.P.), the Singularity Professor Research Project of KAIST (to S.P.), the Institute of Information \& communications Technology Planning \& Evaluation (IITP) (RS-2023-00233251, RS-2019-II190075 to S.W.L.) and the Ministry of Science \& ICT(RS-2024-00436680 to S.W.L.).

\appendix
{
\small

\bibliographystyle{unsrt}
\bibliography{references}
}

\newpage
\input{section/appendix.tex}

\end{document}

%% file: section/introduction.tex
\section{Introduction}

The brain refines its network structure and synaptic connections even before birth, without exposure to sensory stimuli \cite{galli1988, ackman2012, anton2019, martini2021}. In the early developmental stages, the spontaneous neuronal activity that appears in various brain regions is considered to play a critical role during the development of neuronal circuits by pruning neural wiring and adjusting synaptic plasticity \cite{blanquie2017,katz1996,kilb2011,yamamoto2012}. If this activity is disrupted during the developmental stages, the outcome can be long-lasting neuronal deficits \cite{luhmann2016, heck2008, huberman2006}. Computational studies suggest that such a refined network structure enables certain crucial functions of the brain, such as initializing function \cite{kim2021, baek2021, cheon2022, lee2023} and structure \cite{kim2020, albert2008}. These experimental and theoretical studies commonly indicate that spontaneous, random neuronal activity plays a critical role in the development of the biological neural network before data are encountered by the network. However, the detailed mechanism of how these prenatal processes contribute to learning after birth, i.e., with subsequent sensory stimuli, remains elusive.

At the synaptic level, learning can be defined as the process by which the brain adjusts the strength of synaptic connections between neurons to optimize the network for a specific task \cite{gerstner1996, kitazawa1998, keller2009}. The synaptic weights of each neuron can change to minimize the error between the expected and the actual output of a task, often referred to as the credit assignment problem \cite{minsky1961, richards2019, lillicrap2020}. However, in general, it is not well known how individual neurons modify these synaptic connections and thus achieve a network goal under a condition in which numerous neurons are linked in multiple layers. In other words, how neurons can estimate errors to modify their synaptic connections during learning remains unknown.

In machine learning, backpropagation algorithms have successfully addressed this issue – even in deep neural networks \cite{rumelhart1986, krizhevsky2012, lecun2015}. Backpropagation can provide feedback with regard to forward errors through the symmetric copying of forward weights via a backward process. During this process, a structural constraint, i.e., symmetric forward and backward weights, is necessary to assign proper error values to individual neurons \cite{liao2016, saxe2013, kording2001}. However, this process appears to be biologically implausible due to the weight transport problem \cite{grossberg1987, crick1989, lillicrap2016, lillicrap2020}, in which individual neurons must somehow be aware of the exact synaptic connections of their downstream layers to update their weights, a state considered to be practically impossible in a biological brain.

An alternative algorithm, feedback alignment, achieves successful network training even without weight transport by employing fixed random feedback pathways \cite{lillicrap2016}. This study shows that a network can align its weights to synaptic feedback during data training, and this simple process enables error backpropagation. It has been shown that soft alignment between forward and backward weights, which can be achieved during learning data, is enough to back-propagate errors. This finding may provide a biologically plausible scenario in which the credit assignment problem can be resolved, yet there is an issue remaining — the process requires massive data learning to develop the structural constraint, and it significantly underperforms compared to backpropagation on challenging tasks \cite{bartunov2018, shervani-tabar2023}. This cannot be addressed even with currently known advanced learning rules \cite{kolen1994, akrout2019, nokland2016, amit2019, aceituno2023, max2024}.

This situation is contradictory to the notion that the brain can learn even with very limited experience in the initial stages of life \cite{bulf2011, johanson1979, field1984, bushneil1989}. Thus, the question arises as to how early brains can estimate and assign errors for learning with limited experience. To address this issue, here we focus on the role of spontaneous activity at the prenatal stage in the brain, showing that training random noise, which mimics spontaneous random activity in prenatal brains, is a possible solution; random noise training aligns the forward weights to synaptic feedback, enabling precise credit assignment and fast learning. We also observed that random noise training can pre-regulate the weights and enable robust generalization. Our findings suggest that random noise training may be a core mechanism of prenatal learning in biological brains and that it may provide a simple algorithm for the preconditioning of artificial neural networks for fast and robust learning without the weight transport process.

%% file: section/preliminaries.tex
\section{Preliminaries}

Biological and artificial neural networks have different structures and functionalities, but they share certain factors in common, such that information is processed through hierarchical layers of neurons with a nonlinear response function. In the current study, we consider a multi-layer feedforward neural network for pattern classification, $f_\theta: \mathbb{R}^m \rightarrow \mathbb{R}^d$, parameterized by $\theta = \{\mathbf{W}_l, \mathbf{b}_l\}_{l=0}^{L-1}$. It takes input $\mathbf{x} \in \mathbb{R}^m$ and outputs a vector $\mathbf{y} \in \mathbb{R}^d$ with $L$ layers. Through a forward pass, the network computes a hidden layer output by propagating the input through the network layers, as follows:

\vspace{-0.35cm}
\begin{equation}
\label{eq:forward}
    \mathbf{o}_{l+1} = \mathbf{W}_l \mathbf{h}_l + \mathbf{b}_l, \quad \mathbf{h}_{l+1} = \phi(\mathbf{o}_{l+1})
\end{equation}
\vspace{-0.35cm}

, where $\mathbf{W}_l$ is the forward weights, $\mathbf{b}_l$ is the bias vector, and $\phi$ is the nonlinear activation function. In the first layer $l=0$, $\mathbf{h}_l = \mathbf{x}$. We used a rectified linear unit (ReLU) activation function, $\phi(x) = \max(0, x)$. In the last layer $l={L-1}$, we used a softmax function, $\phi_y (x) = \text{softmax}(x) = \{e^{x_i}/\sum_{j=1}^d e^{x_j}\}_{i=1}^d$. Thus, the network outputs a probability distribution over $d$ classes. After the forward pass, the amount of error is calculated by measuring the difference between the network output $f_\theta(\mathbf{x})$ and the target label $\mathbf{y}$. We used the cross-entropy loss \cite{good1952}, which is defined as follows:

\vspace{-0.35cm}
\begin{equation}
\label{eq:loss}
    \mathcal{L}(\theta) = -\frac{1}{N} \sum_{i=1}^N \sum_{j=1}^d \mathbf{y}_{ij} \log f_\theta(\mathbf{x}_i)_j
\end{equation}
\vspace{-0.35cm}

, where $N$ is the number of samples, $d$ is the number of classes, and $\mathbf{y}_{ij}$ is the target label for the $i$-th sample and the $j$-th class. The purpose of learning is to minimize the error $\mathcal{L}(\theta)$. To achieve this, the network parameters $\theta$ are adjusted by assigning credit to the weights that contribute to the error, which is known as the credit assignment problem.

\subsection{Backpropagation and weight transport problem}

To solve the credit assignment problem, backpropagation (BP) \cite{rumelhart1986} computes the gradient of errors with respect to the weights and uses it as a teaching signal to modulate the aforementioned parameters. The gradient is calculated by the chain rule, with propagation from the output layer to the input layer, as follows:

\vspace{-0.35cm}
\begin{equation}
\label{eq:backprop}
\delta_L = \frac{\partial \mathcal{L}}{\partial \mathbf{o}_L} = f_\theta(\mathbf{x}) - \mathbf{y}, \quad \delta_l = \frac{\partial \mathcal{L}}{\partial \mathbf{o}_l} = (\mathbf{W}_l^T \delta_{l+1}) \odot \phi'(\mathbf{o}_l)
\end{equation}
\vspace{-0.35cm}

, where $\delta_l$ is the error signal at layer $l$, $\phi'$ is the derivative of the activation function, and $\odot$ denotes the element-wise product. The weight update rule is given by

\vspace{-0.35cm}
\begin{equation}
\label{eq:weight-update}
\Delta \mathbf{W}_l = -\eta \delta_{l+1} \mathbf{h}_l^T
\end{equation}
\vspace{-0.35cm}

, where $\eta$ is the learning rate. The backpropagation algorithm successfully solves the credit assignment problem, but it requires heavy computation to use the complete information of the synaptic weights of the next layer to update the current weights. Notably, backpropagation is considered as biologically implausible, because it is impossible, in the brain, to transmit the synaptic weights from the next layer to the current layer. This is known as the weight transport problem \cite{grossberg1987, crick1989}.

\subsection{Feedback alignment}

To address the weight transport problem, the idea of feedback alignment (FA) \cite{lillicrap2016} was proposed as a biologically plausible alternative to backpropagation. In feedback alignment, the backward synaptic feedback is replaced with a random, fixed weight matrix $\mathbf{B}_l$ in the feedback path, as follows:

\vspace{-0.35cm}
\begin{equation}
    \label{eq:fa}
    \delta_l = \frac{\partial \mathcal{L}}{\partial \mathbf{o}_l} = (\mathbf{B}_l \delta_{l+1}) \odot \phi'(\mathbf{o}_l).
\end{equation}
\vspace{-0.35cm}

The only difference between backpropagation and feedback alignment is the replacement of the transpose of the forward weights $\mathbf{W}_l$ with the fixed random feedback weights $\mathbf{B}_l$ to calculate the error signal. The fact that the network can learn tasks from error teaching signals that are calculated from random feedback is explained by the observation that the network modifies the forward weights $\mathbf{W}_l$ to match the transpose of the feedback weights $\mathbf{B}_l$ roughly during training. This makes the error teaching signal (\ref{eq:fa}) similar to backpropagation (\ref{eq:backprop}), thus enabling the network to learn the task.

%% file: section/methods.tex
\algnewcommand\algorithmicforeach{\textbf{for each}}
\algdef{S}[FOR]{ForEach}[1]{\algorithmicforeach\ #1\ \algorithmicdo}

\section{Random noise pretraining with feedback alignment}

\begin{algorithm}[ht]
\caption{Random noise pretraining}\label{alg1}
\begin{algorithmic}[1]
\Procedure {Random noise pretraining}{$network$ $f_\theta : \mathbb{R}^m \rightarrow \mathbb{R}^d $}
    \ForEach {epoch}
        \ForEach {batch}
        \State $\mathbf{x} \sim \mathcal{N}(\mu,\sigma^{2}), \; \mathbf{y} \sim \mathcal{U}(0, d-1)$ \Comment{sampling random noise}
        \State $\mathcal{L}(\theta) = \text{Loss}(f_\theta(\mathbf{x}), \mathbf{y})$ \Comment{forward pass, equation (\ref{eq:forward}), (\ref{eq:loss})}
        \State $\delta_L = \frac{\partial \mathcal{L}}{\partial \mathbf{o}_L}$ \Comment{compute error}
        \For {layer $l$ = $L$-1 \textbf{to} $0$}
        \State $\mathbf{W}_{l} = \mathbf{W}_{l} -\eta\,\boldsymbol{\delta}_{l+1}\,\mathbf{h}_{l}^T$ \Comment{update weights, equation (\ref{eq:weight-update})}
        \State $\boldsymbol{\delta}_{l}= (\mathbf{B}_{l}\, \boldsymbol{\delta}_{l+1}) \odot \phi'(\mathbf{o}_l)\ $ \Comment{compute error, equation (\ref{eq:fa})}
        \EndFor
    \EndFor
    \EndFor
\EndProcedure
\end{algorithmic}
\end{algorithm}

During the developmental stage, spontaneous neural activity in the brain plays a critical role in shaping and refining neural circuits. Initially wired immature neural circuits undergo modifications of their connections through the processes of regulated cell formation, apoptosis, and synapse refinement through spontaneous neural activity \cite{martini2021,blanquie2017,katz1996,kilb2011,yamamoto2012}. These pre-sensory activities and development processes are universal across sensory modalities, such as the visual \cite{masland1977, ackman2014}, auditory \cite{tritsch2007, kennedy2012}, and sensorimotor systems \cite{hamburger1966, robinson2000}. We focus here on a few characteristics of spontaneous neural activity in the brain. Spontaneous neural activity is not correlated to external stimuli but can refine and optimize neural circuits, before interaction with the external world can take place.

Here, we propose a type of random training that is inspired by the spontaneous and prenatal neural activity in the brain to pretrain the neural network (Algorithm \ref{alg1}). In every iteration, we sampled random noise inputs $\mathbf{x}$ from a Gaussian distribution $\mathcal{N} (0, I)$ and random labels $\mathbf{y}$ from a discrete uniform distribution $\mathcal{U} (0, N_\text{readout}-1)$, without any correlation. The network $f_\theta$ was initialized with random weights and trained with the feedback alignment algorithm. In this study, we examined the effects of random noise training on the subsequent learning processes in model neural networks to understand the potential benefits of pretraining with random noise in biological brains and whether this strategy is applicable to machine learning algorithms.

%% file: section/results.tex
\section{Results}


\subsection{Weight alignment to synaptic feedback during random noise training}

\begin{figure}[ht]
	\centering
	\includegraphics[width=\textwidth]{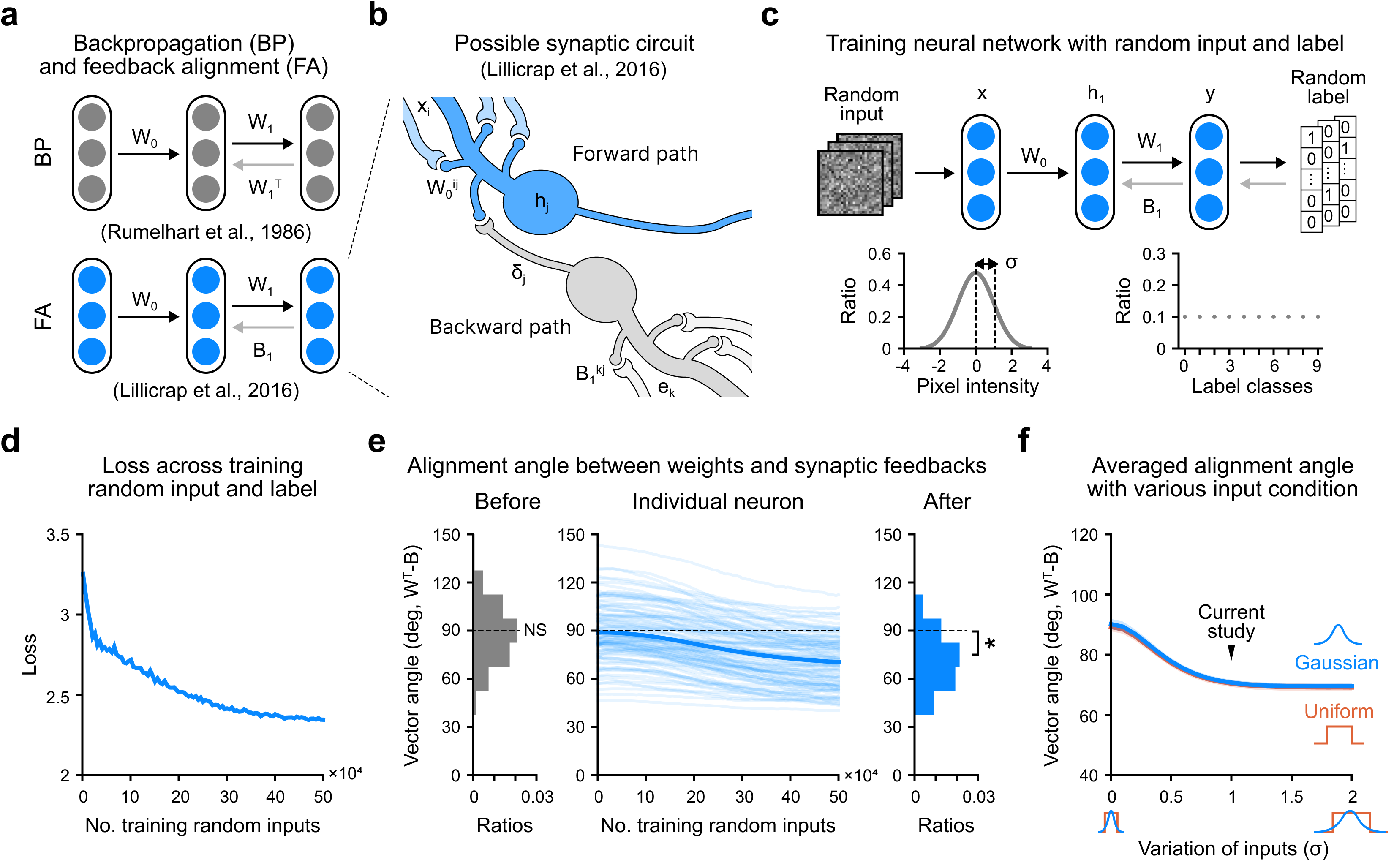}
    \vspace{-0.5cm}
	\caption{Weight alignment to randomly fixed synaptic feedback induced through random noise training.
  (a) Forward and backward pathways of backpropagation and feedback alignment.
  (b) Possible scenario of the feedback alignment algorithm in a biological synaptic circuit.
  (c) Schematic of random training, where the input $\mathbf{x}$ and label $\mathbf{y}$ are randomly sampled and paired in each iteration.
  (d) Cross-entropy loss during random training.
  (e) Alignment angle between forward weights and synaptic feedbacks in the last layer.
  (f) Alignment angle with various random input conditions.}
	\label{fig1}
\end{figure}

To simulate a neural network initially wired by random weights and fixed random synaptic feedback, we adopted a network setting from the feedback alignment algorithm (Figure \ref{fig1}a) in which the weight transport problem can be avoided through the use of fixed random synaptic feedback. Thus, unlike backpropagation, this process is considered possible to exist in biological neural networks with local synaptic connections (Figure \ref{fig1}b). We used a two-layer feedforward neural network with ReLU nonlinearity for classification, $f_\theta: \mathbb{R}^{784} \rightarrow \mathbb{R}^{10}$ with 100 neurons in the hidden layer. By means of random noise training (Algorithm \ref{alg1}), we trained the neural network with random inputs sampled from a Gaussian distribution $\mathcal{N}(0, I)$, with labels also randomly sampled independently (Figure \ref{fig1}c).

We observed that the training loss decreased noticeably during random training, even in the absence of meaningful data and even when $\mathbf{x}$ and $\mathbf{y}$ are randomly paired (Figure \ref{fig1}d). During the random training process, we focused on the alignment between the forward weights and the synaptic feedback. As described in the literature \cite{lillicrap2016}, the alignment of $\mathbf{W}_l$ and $\mathbf{B}_l$, i.e., similarity between $\delta_\text{BP}$ and $\delta_\text{FA}$, is crucial for calculating the error teaching signal precisely. To evaluate the alignment, we used cosine similarity, which is widely used for measuring the distance between two vectors.

\textbf{Definition.} Given the forward weights $\mathbf{W}_l \in \mathbb{R}^{m \times n}$ and backward weights $\mathbf{B}_l \in \mathbb{R}^{n \times m}$, we measured alignment using cosine similarity. We define that $\mathbf{W}_l$ and $\mathbf{B}_l$ as aligned if the angle $\angle (\mathbf{W}_l^T)_i, (\mathbf{B}_l)_i$ is significantly smaller than 90 degrees.

Notably, we found that the weights of neurons are aligned to the corresponding synaptic feedback weights during the random training process (Figure \ref{fig1}e). We also observed that the angle between the forward weights and synaptic feedback of individual neurons in the hidden layer decreased asymptotically during random training. In a randomly initialized network, the alignment angle appeared to be close to 90°, demonstrating that the backward error signal is randomly distributed (Figure \ref{fig1}e, left, alignment angle in an untrained network vs. 90°, $n=100$, one-sample t-test, NS, $P=0.492$). However, after random training, the alignment angle decreased significantly, implying that the backward teaching signal becomes valid to back-propagate errors (Figure \ref{fig1}e, right, alignment angle in an untrained network vs. a randomly trained network, $n=100$, two-sample t-test, $^*P<0.001$). We confirmed that this is not simply due to input bias under a particular condition but is reproduced robustly with various input conditions (Figure \ref{fig1}f). These results suggest that neural networks can pre-learn how to back-propagate errors through random noise training.


\subsection{Pretraining random noise enables fast learning during subsequent data training}

\begin{figure}[ht]
	\centering
	\includegraphics[width=\textwidth]{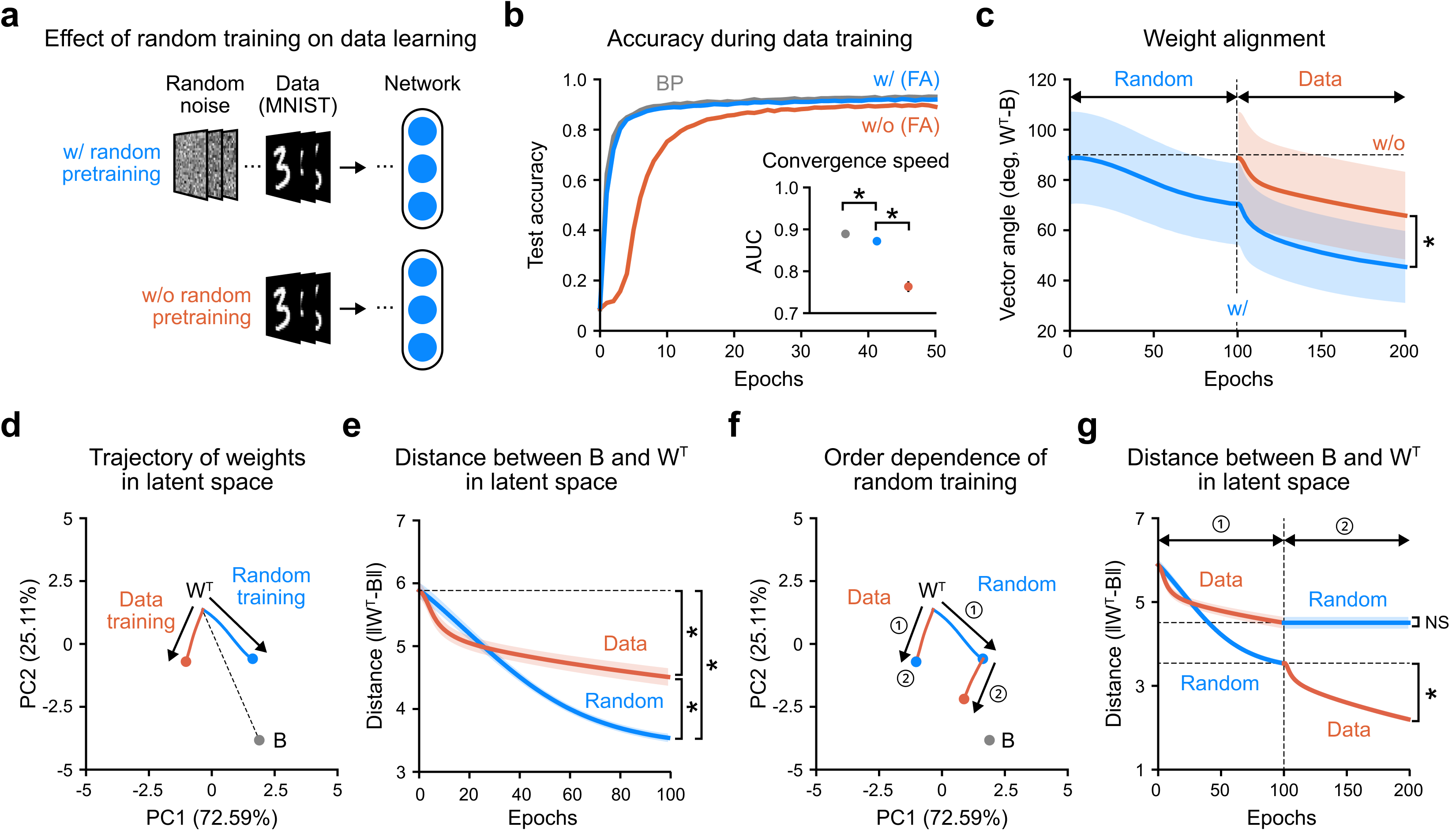}
     \vspace{-0.5cm}
	\caption{Effect of random noise pretraining on subsequent data training.
  (a) Design of the MNIST classification task to investigate the effect of random training.
  (b) Test accuracy during the training process, where the inset demonstrates the convergence speed of each training method, calculated by the AUC of the test accuracy.
  (c) Alignment angle between weights and synaptic feedback across random training and data training.
  (d) Trajectory of weights ($\mathbf{W}_1$) toward synaptic feedback ($\mathbf{B}_1$) in latent space obtained by PCA for random and data training.
  (e) Distance between the weights ($\mathbf{W}_1$) and the synaptic feedback ($\mathbf{B}_1$).
  (f) Order dependence of the trajectory of the weights ($\mathbf{W}_1$).
  (g) Distance between the weights ($\mathbf{W}_1$) and the synaptic feedback ($\mathbf{B}_1$) for different orders of random and data trainings.
  }
	\label{fig2}
\end{figure}

Next, we compared networks with and without random pretraining in terms of subsequent data training outcomes (Figure \ref{fig2}a). We trained the networks using a subset of the MNIST dataset \cite{deng2012}, a widely used benchmark for image classification. We found that a randomly pretrained network learns the data more quickly and achieves higher accuracy compared to a network that is not randomly pretrained (Figure \ref{fig2}b). To quantify the speed of learning, we calculated the area under the curve (AUC) of the test accuracy and found that the convergence of the randomly pretrained network is significantly faster than that in the network without random pretraining (Figure \ref{fig2}b, inset, w/o vs. w/ random pretraining (FA), $n_\text{net}=10$, t-test, $^*P<0.001$). Notably, the convergence speed of the randomly pretrained network appeared comparable to that of the network trained with backpropagation (Figure \ref{fig2}b, inset, w/ random pretraining (FA) vs. BP, $n_\text{net}=10$, t-test, $^*P<0.001$). We also observed that the weight alignment gap between untrained and randomly pretrained networks is maintained during data training (Figure \ref{fig2}c). As a result, at the end of the data training step, the alignment angle of the randomly trained network was significantly smaller than that of the untrained network (Figure \ref{fig2}c, w/o vs. w/ random pretraining, $n=100$, t-test, $^*P<0.001$). This result suggests that a combination of random pretraining and subsequent data training can enhance the weight alignment, which leads to more precise error teaching.

To understand the weight update dynamics by random and data training, we visualized the trajectory of weights in latent space as obtained by a principal component analysis (PCA) \cite{wold1987} (Figure \ref{fig2}d). We conducted PCA on the weights of the last layer ($\mathbf{W}_1$) for the random and data training conditions. First, we confirmed that in both random and data training, the weights become closer to synaptic feedback (Figure \ref{fig2}e, untrained vs. data training, $n_\text{net}=10$, t-test, $^*P<0.001$; untrained vs. random training, $n_\text{net}=10$, t-test, $^*P<0.001$; random vs. data training, $n_\text{net}=10$, t-test, $^*P<0.001$). Notably, we observed that the updated trajectory of weights by random training and data training have different directions in the principal component space and that the effects of random training depend on the order of the random and data training (Figure \ref{fig2}f, g) — the enhancement of weight alignment was more significant when data training was performed after random training (Figure \ref{fig2}g, random training vs. random and data training, $n_\text{net}=10$, t-test, $^*P<0.001$) compared to when training is done in a reversed order (Figure \ref{fig2}g, data training vs. data and random training, $n_\text{net}=10$, t-test, NS, $P=0.999$). Particularly, when we trained the network with data first, subsequent random training could not move the weights; thus, the weights did not become closer to synaptic feedback. This result suggests that weight alignment by random noise pretraining cannot be replaced by data training and that it is crucial to perform random training prior to data training.

\begin{figure}[ht]
  \centering
  \includegraphics[width=\textwidth]{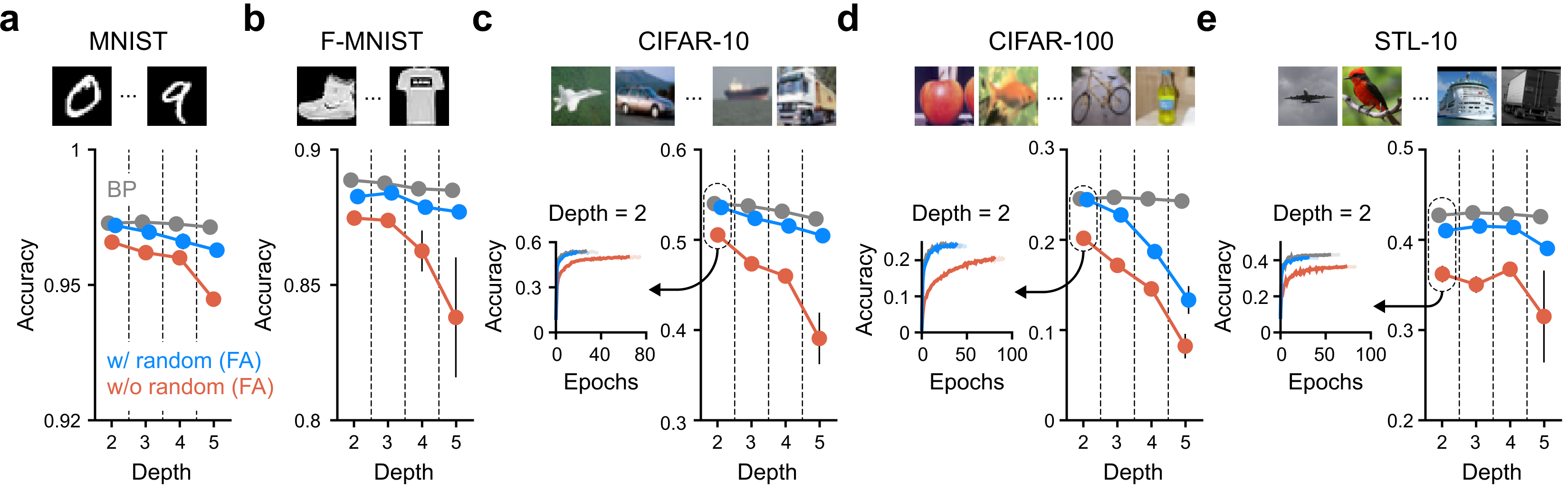}
  \vspace{-0.5cm}
  \caption{Comparison of model performance across different image datasets and network depths.
  (a-e) Final accuracy after convergence. Experiments were conducted with networks of varying depths on different tasks: (a) MNIST, (b) Fashion-MNIST, (c) CIFAR-10, (d) CIFAR-100, and (e) STL-10.
  }
  \label{fig3}
\end{figure}

\begin{table}[htbp]
\footnotesize
\caption{Performance of the two-layer model for each dataset (MNIST, Fashion-MNIST, CIFAR-10, CIFAR-100, STL-10). Each performance value (\%) is presented as the mean $\pm$ standard deviation from three trials. Extended results with various model depths can be found in the Appendix.}
\label{table1}
\centering \begin{tabular}{c c c c c c c}
\toprule
{} & {} & MNIST & F-MNIST & CIFAR-10 & CIFAR-100 & STL-10 \\
\midrule
\multicolumn{2}{c}{BP} & $97.82 \pm 0.03$ & $88.87 \pm 0.03$ & $54.01 \pm 0.20$ & $24.55 \pm 0.10$ & $42.72 \pm 0.20$ \\
\midrule
\multirow{3}{*}{FA} & w/o & $97.26 \pm 0.07$ & $87.47 \pm 0.25$ & $50.54 \pm 0.22$ & $20.17 \pm 0.30$ & $36.21 \pm 0.91$ \\
& w/ & $97.76 \pm 0.07$ & $88.26 \pm 0.07$ & $53.58 \pm 0.12$ & $24.45 \pm 0.10$ & $41.01 \pm 0.16$ \\
\cmidrule{2-7}
& {$\Delta$ACC} & $\blacktriangle 0.49 \pm 0.06$ & $\blacktriangle 0.79 \pm 0.31$ & $\blacktriangle 3.04 \pm 0.20$ & $\blacktriangle 4.28 \pm 0.38$ & $\blacktriangle 4.81 \pm 0.86$ \\
\bottomrule
\end{tabular}
\end{table}

Next, we further investigated the model's classification performance across various image datasets and network depths (Figure \ref{fig3}, Table \ref{table1}). In earlier experiments with two-layer networks and MNIST, we showed that random noise pretraining enhances both the learning speed and accuracy of networks to levels comparable with backpropagation. We extended these experiments to networks of varying depths (Figure \ref{fig3}a) and confirmed that the benefits of random noise pretraining generalize to deeper networks. Additionally, we evaluated performance across various image datasets, including Fashion-MNIST \cite{xiao2017} (Figure \ref{fig3}b), CIFAR-10 \cite{krizhevsky2009} (Figure \ref{fig3}c), CIFAR-100 \cite{krizhevsky2009} (Figure \ref{fig3}d), and STL-10 \cite{coates2011} (Figure \ref{fig3}e). We found that random noise pretraining significantly narrows the performance gap between feedback alignment and backpropagation across different datasets and depths. These results suggest that pretraining with random noise can serve as a general strategy for improving the performance of neural networks trained with feedback alignment algorithms, making it comparable to backpropagation.


\subsection{Pre-regularization by random noise training enables robust generalization}

\begin{figure}[ht]
	\centering
	\includegraphics[width=\textwidth]{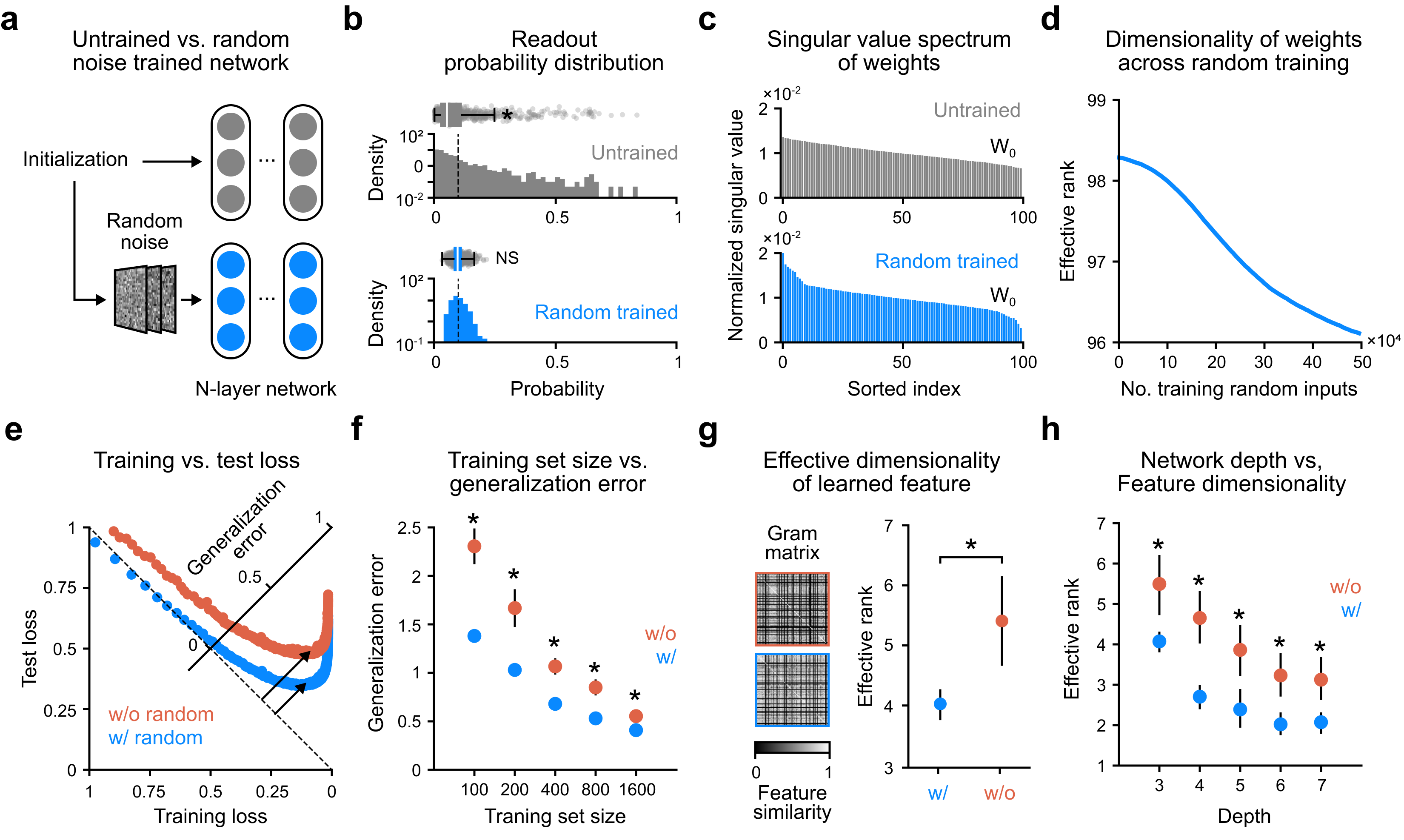}
    \vspace{-0.5cm}
	\caption{Pre-regularization by random noise training enhances generalization
	(a) Untrained network and pre-regularized network through random noise training.
	(b) Distribution of the readout probability.
	(c) Singular value spectrum of the forward weights.
	(d) Effective rank of forward weights during random noise training.
	(e) Generalization error between the training error and test error (training set size: 1600, network depth: 3).
	(f) Generalization error for various training set sizes (network depth: 3).
	(g) Effective dimensionality of the Gram matrix, the cosine similarity of feature vectors across neurons (training set size: 1600, network depth: 3).
	(h) Effective dimensionality of the Gram matrix for various network depths (training set size: 1600).
	}
	\label{fig4}
\end{figure}

Next, we compared the difference between an untrained network and a randomly trained network in terms of their activation and weight (Figure \ref{fig4}a). First, we found that the readout probability of the untrained network is distributed over a wide range (Figure \ref{fig4}b, top, untrained vs. chance level, $n=1000$, Wilcoxon signed-rank test, $^*P<0.001$), whereas that of the randomly trained network is well regularized, close to the chance level (Figure \ref{fig4}b, bottom, random noise trained vs. chance level, $n=1000$, Wilcoxon signed-rank test, NS, $P=0.096$). We also observed that the singular value spectrum of forward weights changes significantly by random training (Figure \ref{fig4}c) such that a small portion of singular values become dominant in the randomly trained network. To measure the effective dimensionality of the weights quantitatively, we used the effective rank of the weights.

\textbf{Definition.} Given a matrix $\mathbf{A} \in \mathbb{R}^{m \times n}$ is decomposed into $\mathbf{A} = U \Sigma V^T$ by singular value decomposition (SVD), the singular values are $\{\sigma_i\}_{i=1}^{min(m, n)}$ sorted in descending order. The effective rank $\rho$ is defined as the Shannon entropy of the normalized singular values, $\rho = -\sum_i {\bar{\sigma}_i}\log {\bar{\sigma_i}}$, where $\bar{\sigma}_i = \sigma_i / \sum_i \sigma_i$. Without loss of generality, we used the effective rank as the exponential of $\rho$ \cite{roy2007}.

We observed that the effective rank of forward weights decreased significantly during random training (Figure \ref{fig4}d), implying that random training regularizes the weights initially and predisposes the network to learn simple solutions of a low rank. Given the notion that low-rank solutions show better generalization performance outcomes, we hypothesized that this pre-regularization by random training enables robust generalization during subsequent data training by inducing low-rank solutions.

To test the generalization ability of the network, we measured the gap between the training error and the test error during subsequent data training. We observed that the generalization error was noticeably lower in a randomly pretrained network compared to a network trained solely on the data (Figure \ref{fig4}e) and that this tendency was maintained with variations of the training set size (Figure \ref{fig4}f, w/o vs. w/ random pretraining, $n_\text{net}=10$, t-test, $^*P<0.001$). This result suggests that pre-regularization by random pretraining can enable robust generalization during subsequent data training.

Next, we compared the representation of learned features in a randomly pretrained network and a network trained without random pretraining. We used the Gram matrix, defined as the cosine similarity of feature vectors across neurons. Notably, we found that the effective rank of the Gram matrix was significantly lower in a randomly pretrained network compared to an untrained network after subsequent data training (Figure \ref{fig4}g, w/o vs. w/ random pretraining, $n_\text{net}=10$, t-test, $^*P<0.001$) and that this tendency was maintained regardless of the network depth (Figure \ref{fig4}h, w/o vs. w/ random pretraining, $n_\text{net}=10$, t-test, $^*P<0.001$). This finding suggests that pre-regularization by random training can enable networks to learn simpler solutions, leading to better generalization performance during subsequent data training.

\begin{figure}[ht]
	\centering
	\includegraphics[width=\textwidth]{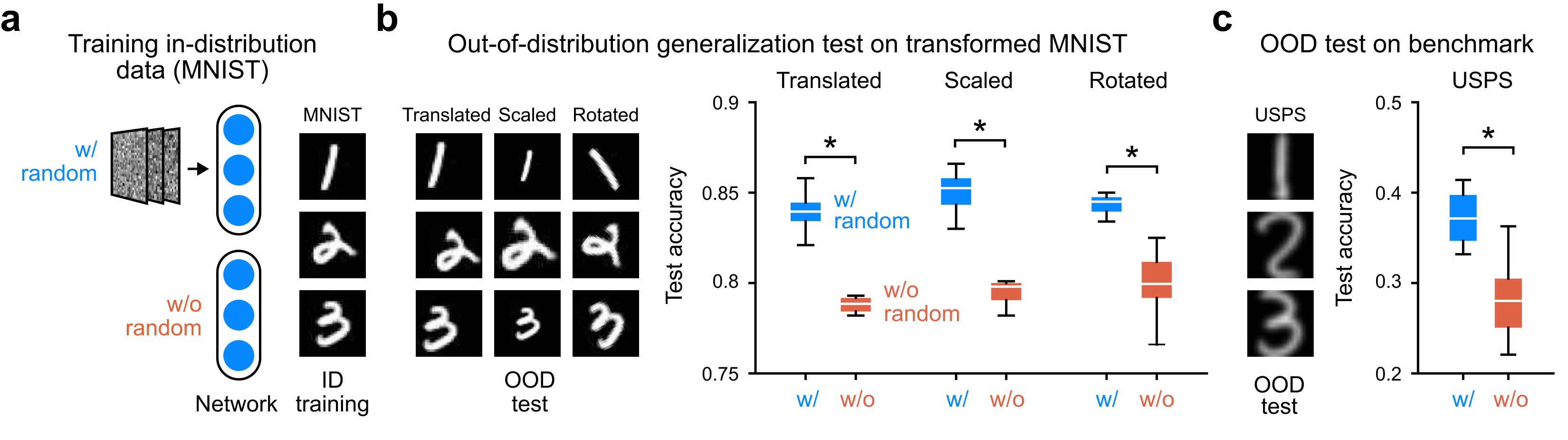}
    \vspace{-0.5cm}
	\caption{Robust generalization of "out-of-distribution" tasks in randomly pretrained networks.
	(a) Training in-distribution data (MNIST) in untrained and randomly pretrained networks.
	(b) Out-of-distribution generalization tests on transformed MNIST.
	(c) Out-of-distribution generalization tests on USPS dataset.
	}
	\label{fig5}
\end{figure}

We also tested the generalization performance of the networks for "out-of-distribution" tasks by training the network with the MNIST dataset and testing it with various out-of-distribution tasks (Figure \ref{fig5}a). First, we generated a MNIST dataset of translated, rotated, and scaled images and then used these images as out-of-distribution tasks (Figure \ref{fig5}b, left). We observed that a randomly pretrained network showed significantly higher test accuracy on out-of-distribution tasks than a network trained without random pretraining (Figure \ref{fig5}b, right, w/o vs. w/ random pretraining, $n_\text{net}=10$, t-test, $^*P<0.001$). We also observed that the randomly pretrained network showed higher test accuracy on the USPS dataset \cite{hull1994}, which is a widely used benchmark dataset for out-of-distribution tasks (Figure \ref{fig5}c, w/o vs. w/ random pretraining, $n_\text{net}=10$, t-test, $^*P<0.001$). This result suggests that pre-regularization by random pretraining enables robust out-of-distribution generalization during subsequent data training.


\subsection{Task-agnostic fast learning for various tasks by a network pretrained with random noise}

\begin{figure}[ht]
	\centering
	\includegraphics[width=\textwidth]{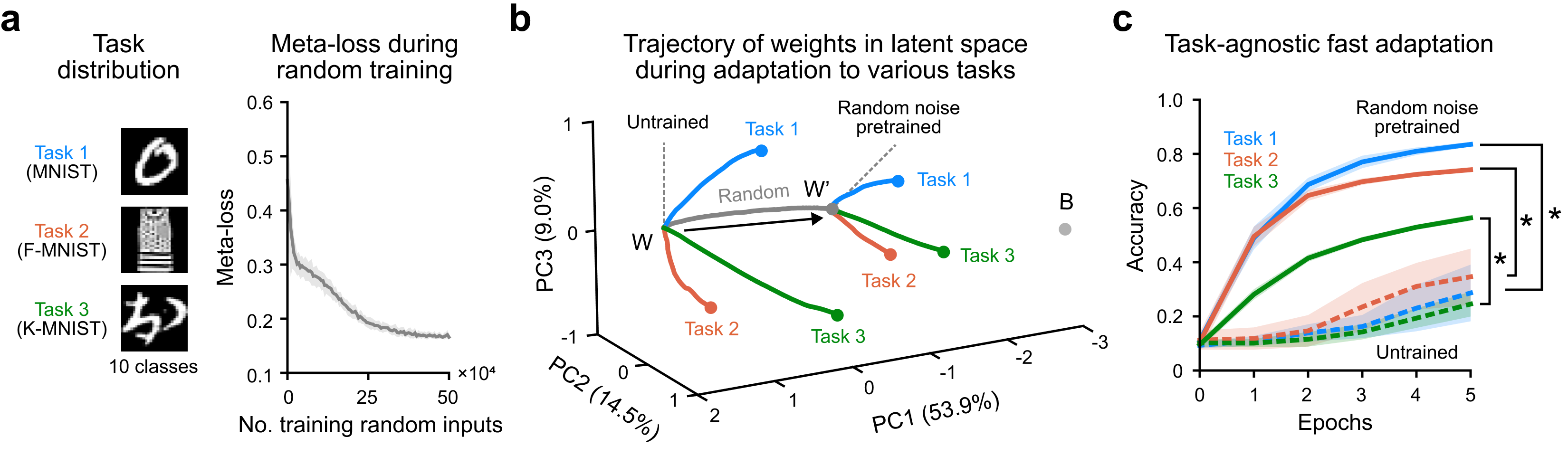}
    \vspace{-0.5cm}
	\caption{Task-agnostic fast learning for various tasks in randomly pretrained networks.
	(a) Three tasks used to test the task-agnostic property of random training, showing the meta-loss during the random training process. The meta-loss is calculated from the sum of the losses measured during adaptation to each task.
	(b) Trajectory of weights in the latent space for adaptation to each task of an untrained network and a randomly pretrained network.
	(c) Adaptation to each task of an untrained network and a randomly pretrained network.
	}
	\label{fig6}
\end{figure}

Lastly, we examined whether random training is generally beneficial for various tasks. We compared the task adaptation capacity of an untrained network and a randomly pretrained network on three tasks: (1) MNIST classification \cite{deng2012}, (2) Fashion-MNIST \cite{xiao2017}, (3) Kuzushiji-MNIST \cite{clanuwat2018} (Figure \ref{fig6}a, left). To measure the ability of fast adaptation to various tasks quantitatively, we computed the meta-loss, as suggested in a previous study of meta-learning.

\textbf{Definition.} Given the task distribution $\mathcal{T} \in \{\mathcal{T}_i\}_{i=1}^{n}$, the meta-loss of network $f_{\theta}$ is defined as $\mathcal{L}_{\text{meta}}(\theta) = \sum_{\mathcal{T}_i \in \mathcal{T}} \mathcal{L}_{\mathcal{T}_i}(\theta_i^{'})$, where $\mathcal{L}_{\mathcal{T}_i}(\theta_i^{'})$ denotes the loss of the task $\mathcal{T}_i$ and $\theta_i^{'}$ is the adapted parameter for $\mathcal{T}_i$ \cite{finn2017}.

We observed that the meta-loss decreased gradually during the random training process (Figure \ref{fig6}a, right). Considering that the training was solely performed with random inputs and labels on the three tasks to measure the meta-loss, this result suggests that networks can learn how to adapt to various tasks without any task-specific data. Next, we trained the untrained networks and randomly pretrained networks on each task separately. We conducted PCA on the weights of the last layer ($\mathbf{W}_2$) to visualize the trajectory of weights in latent space during the adaptation to each task. We observed that the trajectory of weights during random noise training moves closer to synaptic feedback ($\mathbf{B}_2$), which makes the adaptation to each task more efficient (Figure \ref{fig6}b). This suggests that random training is task-agnostic but provides efficient and fast learning in subsequent learning. Lastly, we compared the adaptation to each task in an untrained network and a randomly pretrained network. We observed that the randomly trained network showed remarkably fast adaptation to each task compared to the untrained network (Figure \ref{fig6}c, w/o vs. w/ random pretraining, $n_\text{net}=10$, t-test, $^*P<0.001$). These results highlight the task-agnostic property of random training, which enables networks to quickly adapt to various tasks.

%% file: section/discussion.tex
\section{Discussion}

We showed that random noise pretraining enables neural networks to learn quickly and robustly without weight transport. This finding bridges the gap between a biologically plausible learning mechanism and the conventional backpropagation algorithm, as the symmetry of forward and backward weights can easily be achieved by random noise pretraining. Moreover, the results here provide new insight into the advantage of random training as a means of preconditioning a network for robust generalization.

\textbf{Error-backpropagation without weight transport.} Early work in neuroscience identified basic learning rules, such as Hebbian learning \cite{hebb2005} and spike-timing-dependent plasticity (STDP) \cite{froemke2002, dan2004, caporale2008}. Although these rules have been experimentally observed and are thus biologically plausible, they are not sufficient to explain the brain's remarkable learning ability thoroughly \cite{seung2003, dong2023}. On the other hand, while the backpropagation algorithms used in artificial neural networks have shown impressive learning capabilities, they are considered biologically implausible due to the weight transport problem \cite{lillicrap2020}. Our results provide a new perspective on this issue, bridging the gap between these training rules. We showed that symmetry among forward and backward weights, which is necessary to back-propagate errors, can be readily developed by learning random noise, similar to that during the brain's prenatal stage. Our findings suggest a probable scenario for significantly narrowing the performance gap between previously suggested biologically plausible learning rules and backpropagation.

Recent studies explored noise as a biologically plausible mechanism to enhance learning efficiency without the need for weight transport. For instance, the weight mirror algorithm \cite{akrout2019} uses noisy firing to align feedback weights with forward weights. Similarly, phaseless alignment learning \cite{max2024} leverages layer-wise noise as an additional information carrier to achieve weight alignment. While these approaches have been reported to outperform traditional feedback alignment algorithm, merely incorporating noise into existing feedback alignment algorithms yields no improvement in learning performance \cite{max2024}. In contrast, our results indicate that exposure to random noise "before encountering real data" significantly enhances vanilla feedback alignment. Unlike previous approaches, our strategy utilizes random noise for pre-conditioning the network, preparing it to learn more effectively. This aligns with biological observations that neural noise predominates in the early stage of brain development \cite{skoczenski1998}, even prior to exposure to external stimuli. It is important to note that our proposed method is not limited to the feedback alignment algorithm; pretraining with random noise could be beneficial for other algorithms, which we intend to explore in our follow-up studies.

\textbf{Pre-regularization for robust generalization.} We suggest a task- and model-agnostic pretraining strategy that involves simply training the network with random noise. Notably, our results here show that random noise training can enhance the learning efficiency and generalization ability of the network, for which various tricks and techniques have been proposed to improve. We found that pretraining on random noise reduces the effective dimensionality of the weights, facilitating the learning of low-rank solutions for various tasks. Previous studies on generalization have shown that the low-rank bias of neural networks plays a crucial role in their generalization ability, a finding we have confirmed in this study \cite{arora2019, huh2021, baratin2021, zhao2022, natekar2020}. Additionally, our results highlight that random noise pretraining functions as a form of meta-learning, enhancing the network's ability to adapt rapidly to different tasks. In contrast to previous approaches that utilized data from diverse task distributions \cite{finn2017}, our method achieves similar effects by merely training with noise. It is important to note that the straightforward strategy of random noise training can significantly influence the network's learning dynamics - effects that previous machine learning techniques have sought to achieve. This approach may reflect a potential strategy employed by the brain to attain notable generalization capabilities. Furthermore, it suggests a novel pretraining strategy for artificial neural networks.

\textbf{Insights into developmental neuroscience.} Unlike artificial neural networks, the brain is ready to learn before encountering data. In the early developmental stage before eye-opening, spontaneous random activity emerges in the brain, which is considered essential for the normal development of early circuits \cite{galli1988, ackman2012, anton2019, martini2021}. However, the functional advantage of learning from random noise before external sensory inputs remains unclear. Our study provides a plausible scenario that the brain utilizes spontaneous random activity to pre-align the synaptic weights for error learning and pre-regularization of synaptic connections for robust generalization. Specifically, we showed that random training reduces the effective dimensionality of the weights, which can be considered as a form of pruning, as previous neuroscience studies reported that the brain's synaptic connections are pruned substantially during development, particularly dependent on spontaneous activity \cite{martini2021,blanquie2017,katz1996,kilb2011,yamamoto2012}. Despite the fact that the present study is based on model neural networks, the results here are consistent with a range of experimental findings in developmental neuroscience.

%% file: section/limitations_and_broader_impacts.tex
\section{Broader impacts and limitations}

\textbf{Broader impacts.} Feedback alignment algorithm and its advanced modifications without weight transport are motivated by the need to suggest a learning method that is compatible with deep neural networks with biological plausibility. It can be useful particularly when implemented in physical circuits, as nowadays deep learning without weight transport is utilized in neuromorphic chip engineering. Given that backpropagation requires dynamic access to memory due to weight transport, it is not free from the issue of energy inefficiency. Our results are not solely limited to demonstrating the role of biological prenatal learning but can also be extended for more practical purposes; for instance, it is a promising strategy for the preconditioning of neuromorphic chips.

\textbf{Limitations.} Although our study offers a new perspective on pretraining neural networks with random noise, some limitations must be considered. The current study focuses on results using feedforward neural networks with feedback alignment algorithms. Regarding the scalability of the method, further investigation is needed for other architectures, such as convolutional neural networks. Notably, we achieved meaningful results showing that pretraining with random noise can also benefit standard backpropagation, which will be further explored in our follow-up studies, which will emphasize random noise pretraining as a general strategy for neural network training.

%% file: section/appendix.tex
\appendix
\setcounter{table}{0}
\renewcommand{\thetable}{S\arabic{table}}
\setcounter{figure}{0}
\renewcommand{\thefigure}{S\arabic{figure}}

\section{Experimental details and additional results for section 4.1}

\subsection{Network architecture and training details}

\begin{table}[H]
    \caption{Parameters and settings used in the experiment.}
    \label{tablea1}
    \centering
    \begin{tabular}{lc}
        \toprule
        Name                      & Setting            \\
        \midrule
        Dimensions                & {[}784, 100, 10{]} \\
        Activation function       & ReLU               \\
        Number of random noise inputs      & $5 \times 10^5$    \\
        Batch size                & 64                 \\
        Learning rate             & 0.0001             \\
        Optimizer                 & Adam               \\
        \bottomrule
    \end{tabular}
\end{table}

In Section 4.1, we utilized a two-layer feedforward neural network for classification, comprising 100 neurons in the hidden layer and employing the rectified linear unit (ReLU) as the activation function. Detailed hyperparameters related to the network architecture and training are presented in Table \ref{tablea1}. The forward weights were initialized using He initialization \cite{he2015}, which is standard for networks utilizing ReLU activations. Specifically, the weights were sampled from a Gaussian distribution with a mean of zero and a standard deviation of:

\[ \sigma = \sqrt{\frac{2}{n_{\text{in}}}} \]

, where $n_\text{in}$ represents the number of input units for each layer. The biases were initialized to zero. For feedback alignment \cite{lillicrap2016}, the backward weights were randomly initialized from the same distribution as the forward weights. In contrast to backpropagation, these backward weights remained fixed throughout the entire training process and were not updated. This approach allowed us to circumvent weight transport while still facilitating alignment during random noise pretraining and data training.

\begin{figure}[H]
	\centering
	\includegraphics[width=\textwidth]{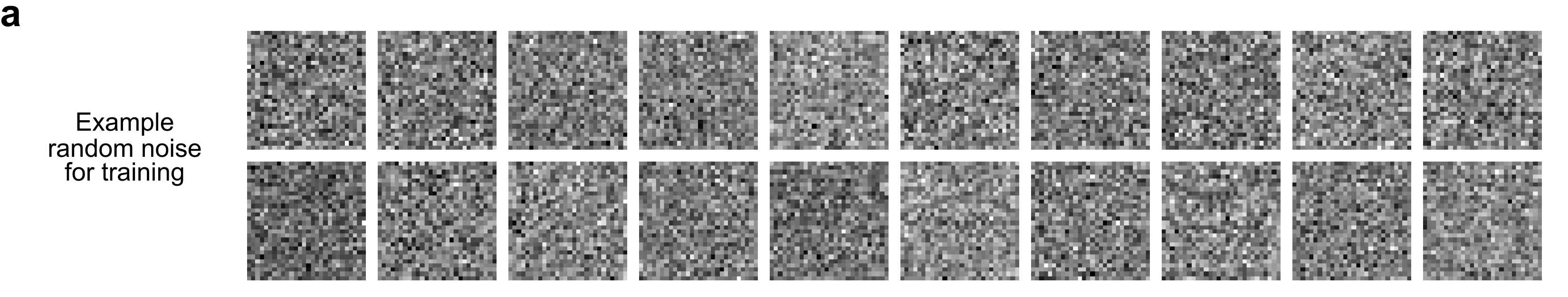}
	\caption{Samples of random noise utilized in the pretraining process. Each pixel value is randomly drawn from a zero-centered Gaussian distribution with a standard deviation of 1. The corresponding label is also randomly assigned from a discrete uniform distribution ranging from 0 to 9.}
	\label{figa1}
\end{figure}

During the random noise pretraining phase, we sampled random input-output pairs at each iteration. Each sampled pair was used only once throughout the training process. Figure \ref{figa1} presents an example of a random noise input sampled from a Gaussian distribution. This method of random noise pretraining was applied consistently across all subsequent experiments described in the main text.

It is important to note that random noise pretraining does not depend on specific conditions for the random input. To further investigate the robustness of this method, we tested weight alignment under various input conditions (Figure 1f). Specifically, we varied the standard deviation of the Gaussian distribution from 0 to 2, with a step size of 0.1. Additionally, we conducted similar tests using uniform distributions. The reported results include the mean and standard deviation of the alignment angle, calculated over ten independent trials.

\newpage
\section{Experimental details and additional results for section 4.2}

\subsection{Network architecture and training details}

\begin{table}[H]
    \caption{Parameters and settings used in the experiment.}
    \label{tablea2}
    \centering
    \begin{tabular}{lc}
        \toprule
        Name                      & Setting            \\
        \midrule
        Number of training data   & $5 \times 10^3$    \\
        Number of test data       & $5 \times 10^3$    \\
        Epochs                    & 100                \\
        Batch size                & 64                 \\
        Learning rate             & 0.0001             \\
        Optimizer                 & Adam               \\
        \bottomrule
    \end{tabular}
\end{table}

\begin{figure}[H]
	\centering
	\includegraphics[width=\textwidth]{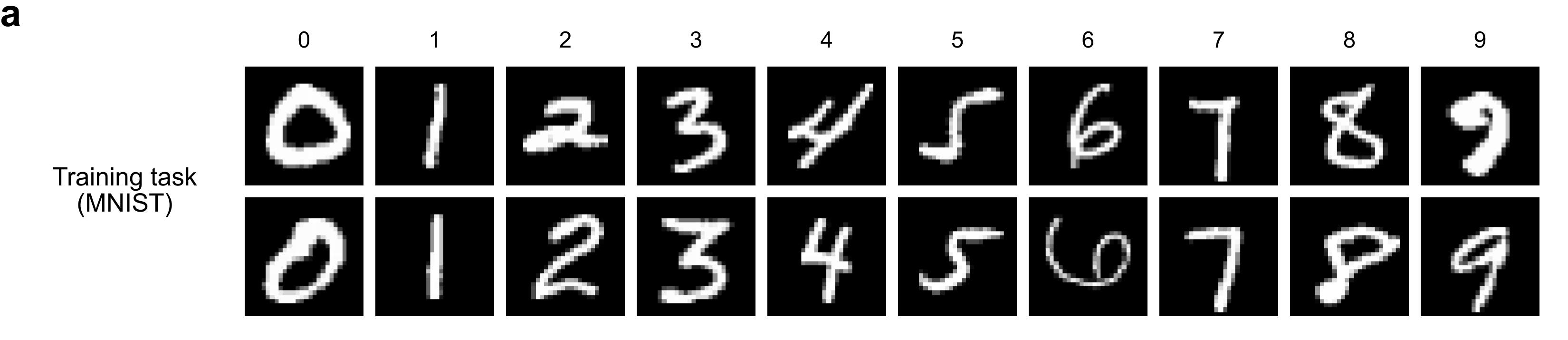}
	\caption{Samples from the MNIST dataset \cite{deng2012} used in the subsequent data training process. The dataset comprises images of handwritten digits across ten classes.}
	\label{figa2}
\end{figure}

In Section 4.2, we examined the effect of random noise pretraining on subsequent data training. The network architecture and hyperparameters were consistent with those used in Section 4.1. Detailed hyperparameters related to the training process can be found in Table \ref{tablea2}. We utilized a subset of the MNIST dataset (Figure \ref{figa2}), consisting of 5000 images for training and an additional 5000 images for testing. The data training was conducted over a total of 100 epochs, ensuring that the total number of inputs encountered during the data training process matched those in the random noise pretraining phase.

Although random noise pretraining does not conform to the conventional definition of an epoch — since it does not involve repeated training on the same inputs — we employed the term "epoch" to facilitate comparison between random noise training and data training along the same axis. Specifically, we divided random noise pretraining into 5000 iterations, analogous to the number of iterations in data training. Thus, one epoch utilized the same number of inputs (along with equivalent training time and computational cost) in both random noise pretraining and data training.

\newpage
\subsection{Training results}

\begin{figure}[H]
	\centering
	\includegraphics[width=0.75\textwidth]{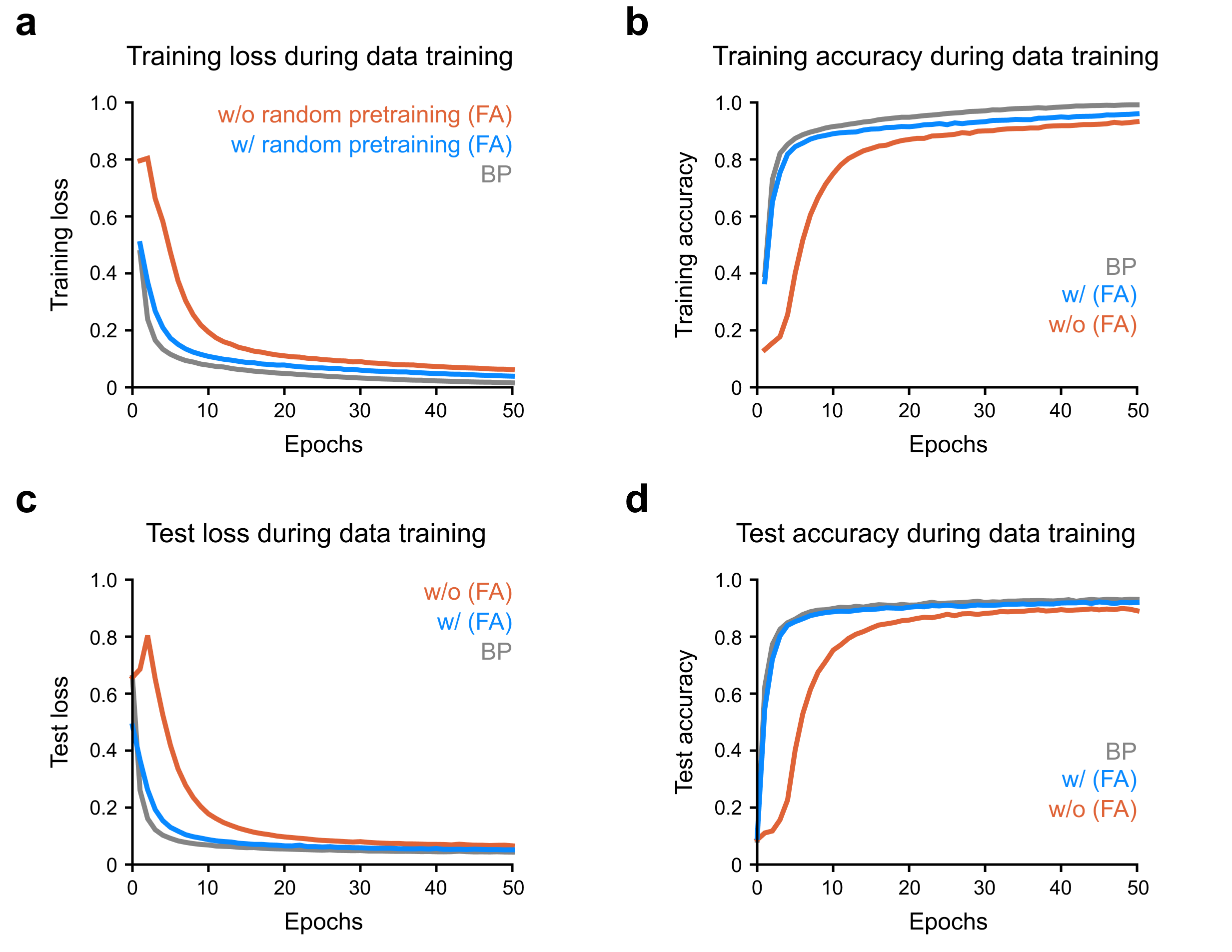}
	\caption{Loss and accuracy outcomes of the network under various training conditions. (a) Training loss. (b) Training accuracy. (c) Test loss. (d) Test accuracy. The blue and orange lines represent the network trained with feedback alignment, with and without random noise training, respectively. The gray lines denote the network trained using backpropagation.}
	\label{figa3}
\end{figure}

In addition to the test accuracy during training presented in Figure 2b, Figure \ref{figa3} displays the loss and accuracy for both the training and test sets. The results indicate that the network pretrained with random noise outperforms the network trained solely on data and achieves a learning efficiency comparable to that of backpropagation \cite{rumelhart1986}.

\newpage
\subsection{Additional results with deeper network}

\begin{table}[H]
    \caption{Parameters and settings used in the experiment.}
    \label{tablea3}
    \centering
    \begin{tabular}{lc}
        \toprule
        Name                      & Setting                 \\
        \midrule
        Dimensions                & {[}784, 100, 100, 10{]} \\
        Activation function       & ReLU                    \\
        Number of random noise inputs        & $5 \times 10^5$         \\
        Number of training data   & $5 \times 10^3$         \\
        Number of test data       & $5 \times 10^3$         \\
        Epochs                    & 100                     \\
        Batch size                & 64                      \\
        Learning rate             & 0.0001                  \\
        Optimizer                 & Adam                   \\
        \bottomrule
    \end{tabular}
\end{table}

\begin{figure}[H]
    \centering
    \includegraphics[width=0.75\textwidth]{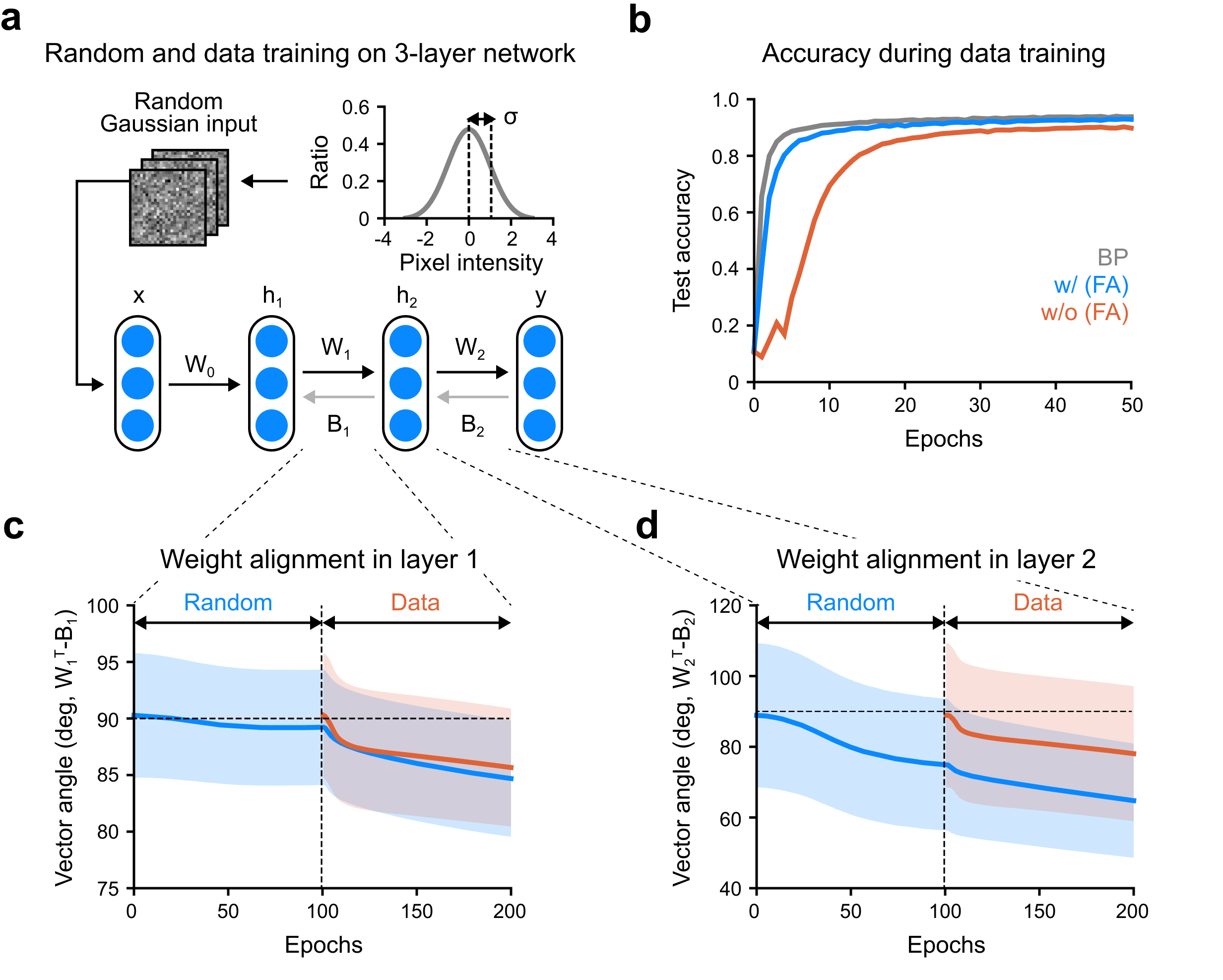}
    \caption{Training with random noise and data in a deeper network. (a) Three-layer network trained sequentially with random noise and data. (b) Test accuracy of the network under various training conditions. (c) Alignment angle of the weights ($\mathbf{W}_1$) and synaptic feedback ($\mathbf{B}_1$) in the second layer of the network. (d) Alignment angle of the weights ($\mathbf{B}_2$) and synaptic feedback ($\mathbf{W}_2$) in the third layer of the network.}
    \label{figa4}
\end{figure}

We also investigated whether weight alignment improves in deeper networks. Specifically, we employed a three-layer feedforward neural network with a hidden layer size of 100 and examined the effect of random noise pretraining (Figure \ref{figa4}a). Detailed hyperparameters related to the network architecture and training can be found in Table \ref{tablea3}. The results indicate that random noise pretraining remains beneficial, achieving faster learning and higher accuracy during subsequent data training (Figure \ref{figa4}b). Additionally, the network pretrained with random noise exhibited enhanced weight alignment, comparable to networks trained solely on data, in both the earlier layer (Figure \ref{figa4}c) and the deeper layer (Figure \ref{figa4}d).

\newpage
\subsection{Additional results with CIFAR-10}

\begin{table}[H]
    \caption{Parameters and settings used in the experiment.}
    \label{tablea4}
    \centering
    \begin{tabular}{lc}
        \toprule
        Name                      & Setting                 \\
        \midrule
        Dimensions                & {[}3072, 100, 10{]}     \\
        Activation function       & ReLU                    \\
        Number of random noise inputs        & $5 \times 10^5$         \\
        Number of training data   & $5 \times 10^3$         \\
        Number of test data       & $5 \times 10^3$         \\
        Epochs                    & 100                     \\
        Batch size                & 64                      \\
        Learning rate             & 0.0001                  \\
        Optimizer                 & Adam                   \\
        \bottomrule
    \end{tabular}
\end{table}

\begin{figure}[H]
	\centering
	\includegraphics[width=\textwidth]{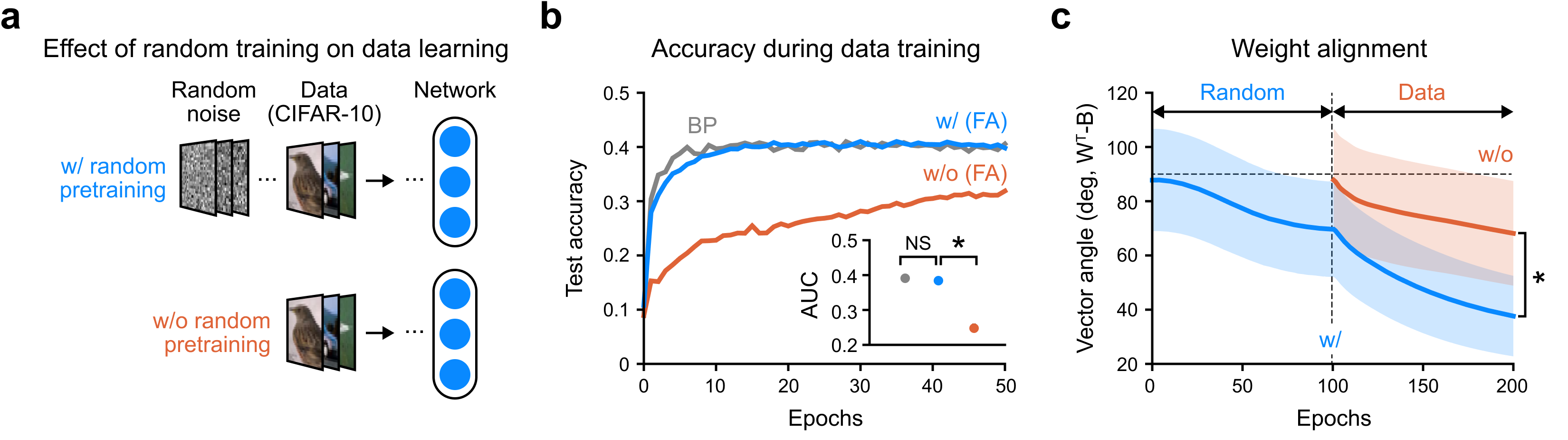}
        \vspace{-0.3cm}
	\caption{Effect of random noise training on subsequent CIFAR-10 \cite{krizhevsky2009} data training.
    (a) Design of the CIFAR-10 classification task.
    (b) Test accuracy during the training process, where the inset demonstrates the convergence speed of each training method, calculated by the AUC of the test accuracy.
    (c) Alignment angle between weights ($\mathbf{W}_1$) and synaptic feedback ($\mathbf{B}_1$) across random training and data training.}
    \vspace{-0.3cm}
	\label{figa5}
\end{figure}

\begin{figure}[H]
	\centering
	\includegraphics[width=\textwidth]{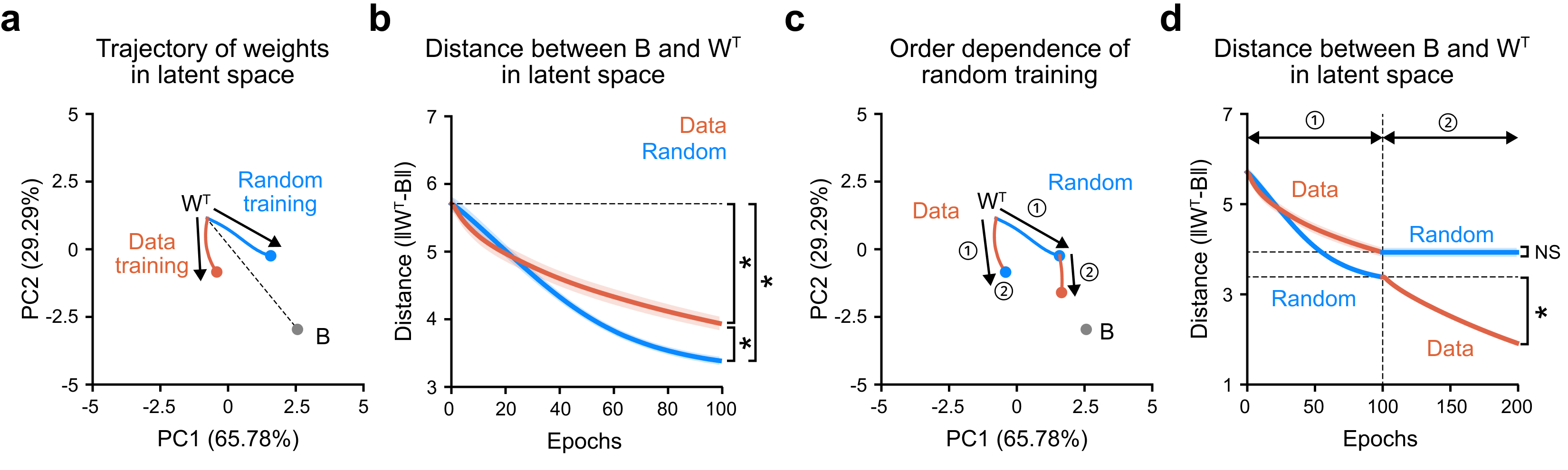}
 \vspace{-0.3cm}
	\caption{Weight update dynamic of random noise training and data training with CIFAR-10.
    (a)	Trajectory of weights ($\mathbf{W}_1$) toward synaptic feedback ($\mathbf{B}_1$) in latent space obtained by PCA for random and data training.
    (b) Distance between the weights ($\mathbf{W}_1$) and the synaptic feedback ($\mathbf{B}_1$).
    (c) Order dependence of the trajectory of the weights ($\mathbf{W}_1$).
    (d) Distance between the weights ($\mathbf{W}_1$) and the synaptic feedback ($\mathbf{B}_1$) for different orders of random and data trainings.}
    \vspace{-0.3cm}
	\label{figa6}
\end{figure}

To extend the results demonstrating the benefits of random noise pretraining on subsequent MNIST training, we conducted a similar experiment using the CIFAR-10 dataset, which contains more naturalistic images across ten different classes of objects and animals (Figure \ref{figa5}, Figure \ref{figa6}). Detailed hyperparameters related to the network architecture and training can be found in Table \ref{tablea4}. These results emphasize that the benefits and characteristics of random noise pretraining, as observed in the main results, are not confined to a simple dataset like MNIST but are applicable to other datasets as well.

\newpage
\subsection{Validation of model performance across various image datasets and network depths}

\begin{table*}[ht]
\caption{Performance of each model with depth variation (2 – 5 layers) for five different datasets (MNIST, Fashion-MNIST, CIFAR-10, CIFAR-100 and STL-10). Each performance value (\%) is presented as the mean $\pm$ standard deviation from three trials.}
\vspace{0.3cm}
\label{tablea5}
\centering
\small
\begin{tabular}{c c c c c c c}
\toprule
& \multicolumn{2}{c}{Method} & {2-layer} & {3-layer} & {4-layer} & {5-layer} \\
\midrule
\multirow{4}{*}{\rotatebox[origin=c]{90}{MNIST}} 
& \multicolumn{2}{c}{BP} & $97.82 \pm 0.03$ & $97.86 \pm 0.04$ & $97.80 \pm 0.16$ & $97.71 \pm 0.18$ \\
\cmidrule{2-7}
& \multirow{2}{*}{FA} & w/o & $97.26 \pm 0.07$ & $96.95 \pm 0.12$ & $96.80 \pm 0.20$ & $95.58 \pm 0.15$ \\ 
& & w/ & $97.76 \pm 0.07$ & $97.56 \pm 0.09$ & $97.29 \pm 0.24$ & $97.03 \pm 0.04$ \\ 
\cmidrule{2-7}
& \multicolumn{2}{c}{$\Delta$ACC} & $\blacktriangle 0.49 \pm 0.06$ & $\blacktriangle 0.61 \pm 0.11$ & $\blacktriangle 0.48 \pm 0.28$ & $\blacktriangle 1.45 \pm 0.19$ \\ 
\midrule

\multirow{4}{*}{\rotatebox[origin=c]{90}{F-MNIST}} 
& \multicolumn{2}{c}{BP} & $88.87 \pm 0.03$ & $88.76 \pm 0.09$ & $88.55 \pm 0.00$ & $88.50 \pm 0.13$ \\
\cmidrule{2-7}
& \multirow{2}{*}{FA} & w/o & $87.47 \pm 0.25$ & $87.38 \pm 0.20$ & $86.25 \pm 0.76$ & $83.80 \pm 2.22$ \\ 
& & w/ & $88.26 \pm 0.07$ & $88.40 \pm 0.02$ & $87.87 \pm 0.13$ & $87.70 \pm 0.21$ \\ 
\cmidrule{2-7}
& \multicolumn{2}{c}{$\Delta$ACC} & $\blacktriangle 0.79 \pm 0.31$ & $\blacktriangle 1.02 \pm 0.19$ & $\blacktriangle 1.62 \pm 0.86$ & $\blacktriangle 3.90 \pm 2.35$ \\ 
\midrule

\multirow{4}{*}{\rotatebox[origin=c]{90}{CIFAR-10}} 
& \multicolumn{2}{c}{BP} & $54.01 \pm 0.20$ & $53.75 \pm 0.22$ & $53.16 \pm 0.20$ & $52.32 \pm 0.03$ \\
\cmidrule{2-7}
& \multirow{2}{*}{FA} & w/o & $50.54 \pm 0.22$ & $47.34 \pm 0.83$ & $45.98 \pm 0.33$ & $39.07 \pm 2.85$ \\ 
& & w/ & $53.58 \pm 0.12$ & $52.38 \pm 0.05$ & $51.54 \pm 0.28$ & $50.46 \pm 0.81$ \\ 
\cmidrule{2-7}
& \multicolumn{2}{c}{$\Delta$ACC} & $\blacktriangle 3.04 \pm 0.20$ & $\blacktriangle 5.04 \pm 0.82$ & $\blacktriangle 5.55 \pm 0.09$ & $\blacktriangle 11.39 \pm 2.48$ \\ 
\midrule

\multirow{4}{*}{\rotatebox[origin=c]{90}{CIFAR-100}} 
& \multicolumn{2}{c}{BP} & $24.55 \pm 0.10$ & $24.71 \pm 0.04$ & $24.54 \pm 0.08$ & $24.29 \pm 0.13$ \\
\cmidrule{2-7}
& \multirow{2}{*}{FA} & w/o & $20.17 \pm 0.30$ & $17.17 \pm 0.54$ & $14.56 \pm 0.13$ & $8.22 \pm 1.36$ \\ 
& & w/ & $24.45 \pm 0.10$ & $22.76 \pm 0.42$ & $18.69 \pm 0.65$ & $13.33 \pm 1.54$ \\ 
\cmidrule{2-7}
& \multicolumn{2}{c}{$\Delta$ACC} & $\blacktriangle 4.28 \pm 0.38$ & $\blacktriangle 5.58 \pm 0.18$ & $\blacktriangle 4.14 \pm 0.52$ & $\blacktriangle 5.11 \pm 1.31$ \\ 
\midrule

\multirow{4}{*}{\rotatebox[origin=c]{90}{STL-10}} 
& \multicolumn{2}{c}{BP} & $42.72 \pm 0.20$ & $43.00 \pm 0.28$ & $42.87 \pm 0.08$ & $42.56 \pm 0.22$ \\
\cmidrule{2-7}
& \multirow{2}{*}{FA} & w/o & $36.21 \pm 0.91$ & $35.04 \pm 0.92$ & $36.74 \pm 0.24$ & $31.50 \pm 5.10$ \\ 
& & w/ & $41.01 \pm 0.16$ & $41.52 \pm 0.15$ & $41.38 \pm 0.14$ & $39.04 \pm 0.59$ \\ 
\cmidrule{2-7}
& \multicolumn{2}{c}{$\Delta$ACC} & $\blacktriangle 4.81 \pm 0.86$ & $\blacktriangle 6.49 \pm 1.03$ & $\blacktriangle 4.64 \pm 0.35$ & $\blacktriangle 7.54 \pm 5.37$ \\ 
\bottomrule
\end{tabular}%
\end{table*}

\begin{figure}[H]
	\centering
	\includegraphics[width=\textwidth]{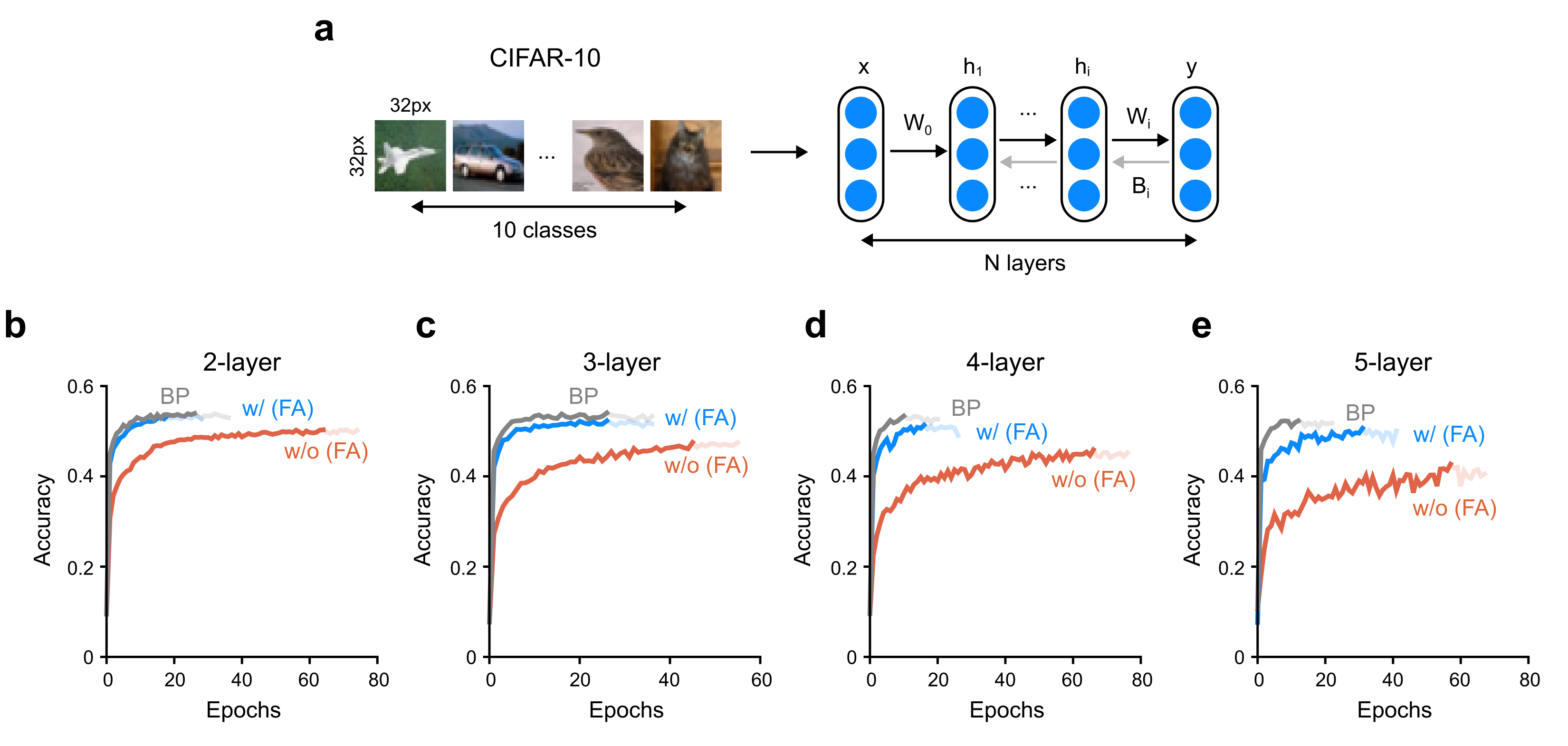}
	\caption{Model performance across different network depths.
 (a-d) Validation accuracy measured at each epoch during data training. Each colored line represents a network trained using either feedback alignment or backpropagation. Experiments were conducted with various depths of MLP: (a) two-layer, (b) three-layer, (c) four-layer, and (d) five-layer.}
	\label{figa7}
\end{figure}

\begin{figure}[H]
	\centering
	\includegraphics[width=\textwidth]{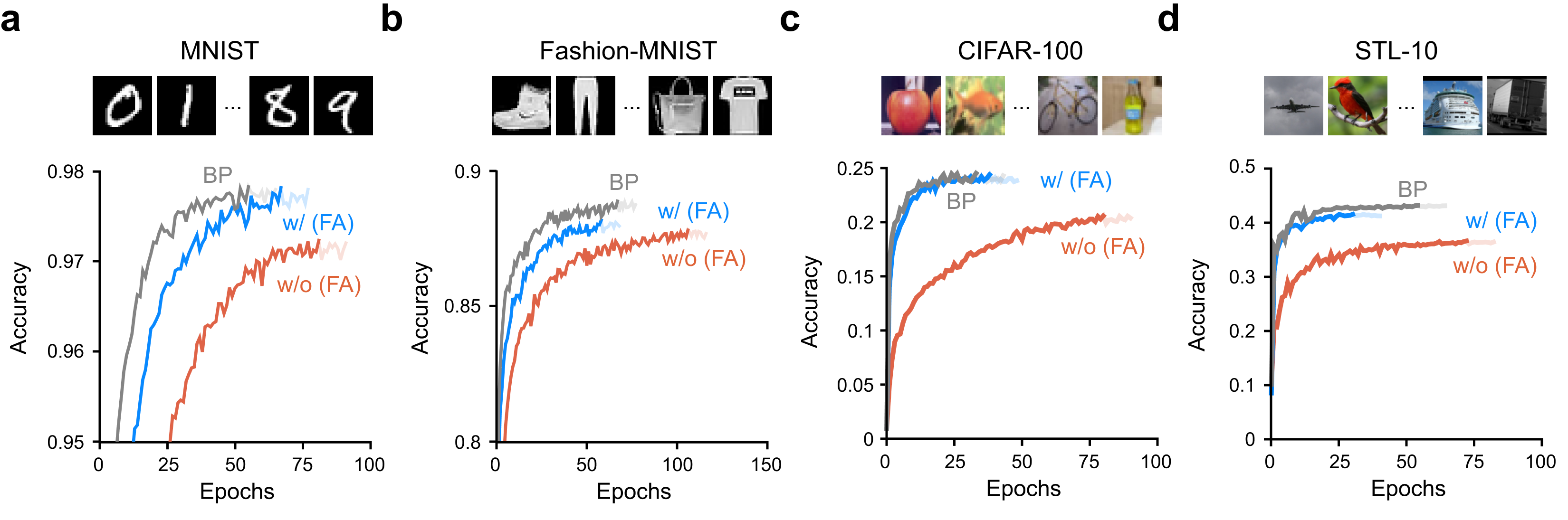}
	\caption{Model performance across different image datasets.
 (a-d) Validation accuracy is measured at each epoch. (a) MNIST, (b) Fashion-MNIST, (c) CIFAR-100, (d) STL-10.}
	\label{figa8}
\end{figure}

We benchmarked the final accuracy of networks trained using baseline feedback alignment (FA, w/o random pretraining), feedback alignment with random noise pretraining (FA, w/ random pretraining: our model), and backpropagation (BP) (Table \ref{tablea5}). Specifically, we evaluated these models at depths ranging from 2 to 5 layers (Figure \ref{figa7}). Furthermore, we compared these models across several complex and large datasets, including MNIST \cite{deng2012}, Fashion-MNIST \cite{xiao2017}, CIFAR-10 \cite{krizhevsky2009}, CIFAR-100 \cite{krizhevsky2009}, and STL-10 \cite{coates2011} (Figure \ref{figa8}). To ensure full convergence during training, we extended the training duration until convergence was confirmed, with validation accuracy showing no further increase (patience: 10 epochs).

We observed that incorporating random noise training at various network depths and across diverse datasets consistently resulted in higher final accuracy, often comparable to that achieved with backpropagation. As the network depth increased, the disparity in final accuracy between models with and without random noise training widened (Figure \ref{figa7}). Importantly, the beneficial impact of random noise training on final accuracy became significantly more pronounced as task complexity increased (Figure \ref{figa8}). For instance, while random noise training improved accuracy by 0.49\% in MNIST, it increased by 0.79\% in Fashion-MNIST, 3.04\% in CIFAR-10, 4.28\% in CIFAR-100, and 4.81\% in STL-10. This gap widened further as the network depth increased.

\newpage

Indeed, previous studies on biologically plausible backpropagation without weight transport often focus on simple network structures and easy datasets. This limitation results in lower learning capacity and presents challenges in scaling up due to biological constraints that preclude weight transport. Through our additional experiments, we have demonstrated that random noise pretraining consistently outperforms baseline feedback alignment, even in deeper networks. However, we also observed that the performance gap with backpropagation widens as the number of layers increases. This challenge partly arises from the difficulty of achieving precise weight alignment in the early layers using feedback alignment without weight transport. While our approach significantly enhances the learning efficiency of feedback alignment algorithms, the issue of the performance gap in deep networks compared to backpropagation remains unresolved.

\newpage
\subsection{Validation of model performance across various training hyperparameters}
\begin{figure}[H]
	\centering
	\includegraphics[width=0.75\textwidth]{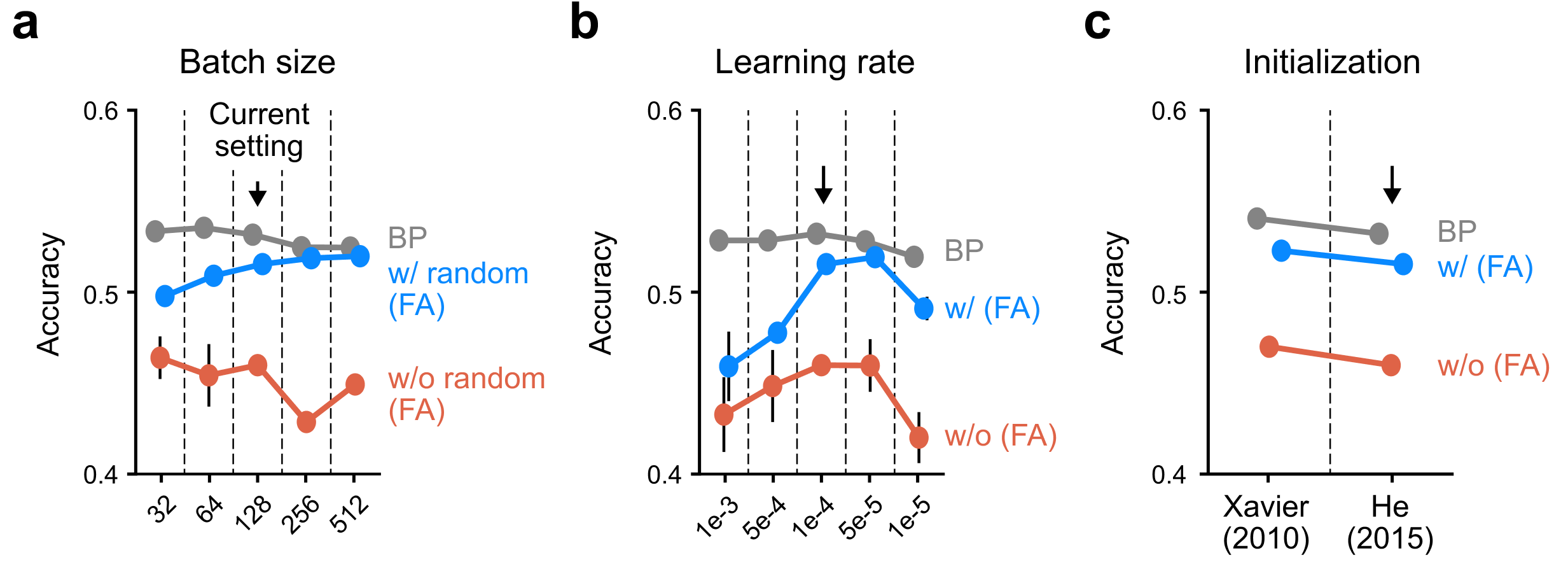}
	\caption{Performance consistency across different hyperparameter settings and initializations.
 (a-c) Accuracy of networks trained with varying hyperparameters: (a) batch size, (b) learning rate, and (c) initialization.}
	\label{figa9}
\end{figure}

Our results are based on hyperparameters typically selected for effective learning in both feedback alignment and backpropagation. We conducted and compared baseline feedback alignment (FA, w/o random pretraining), feedback alignment with random noise pretraining (FA, w/ random pretraining), and backpropagation (BP) under carefully controlled conditions. We controlled hyperparameters such as batch size, learning rate, and initial forward weights to isolate differences attributable to the learning algorithms. In each trial, we initialized a single network randomly and duplicated it into three, starting with identical weights. These networks were then trained with the same hyperparameters but using different learning methods.

Additionally, we confirmed that our results were not specific to the choice of hyperparameters — our findings generalized to other conditions. Further experiments using the CIFAR-10 dataset demonstrated that varying hyperparameters (batch size, learning rate) and initialization within a reasonable range consistently reproduced similar results (Figure \ref{figa9}).

\newpage
\subsection{Model validation with initialization matching forward and backward weights}

\begin{figure}[H]
	\centering
	\includegraphics[width=0.75\textwidth]{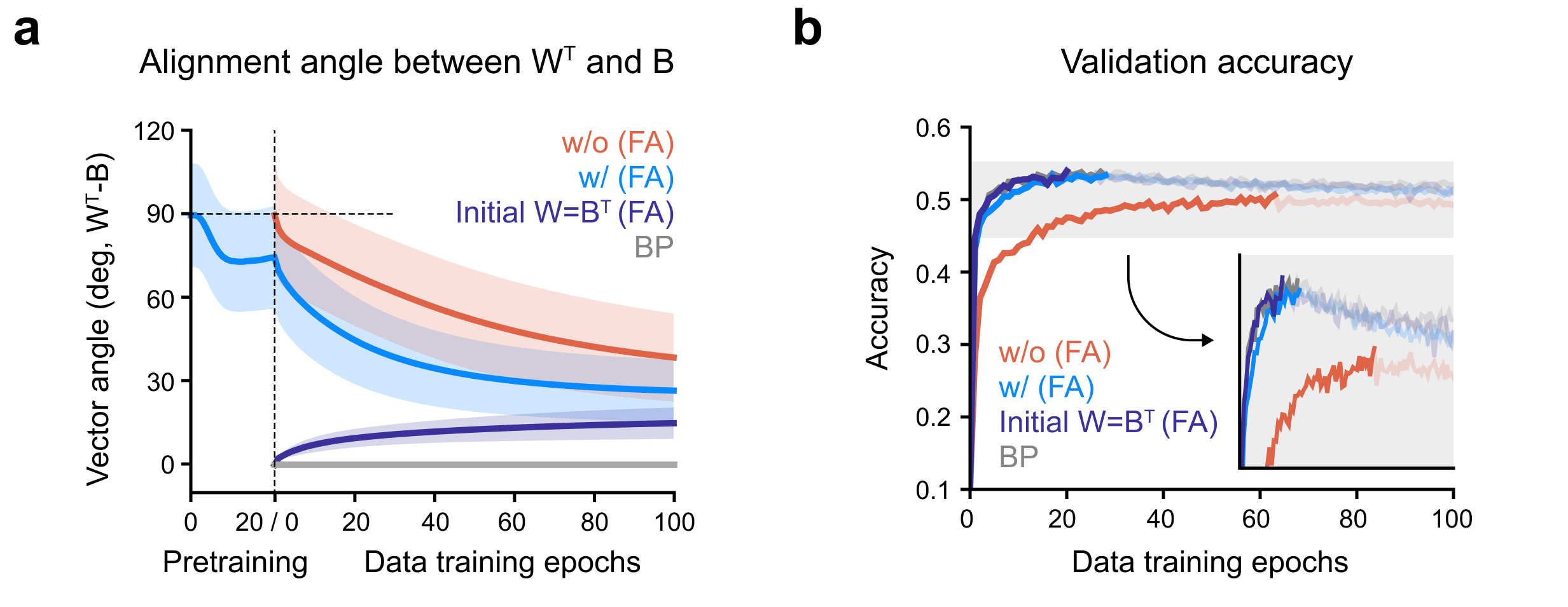}
	\caption{Model validation: Initial alignment of weights ($\mathbf{W}_l$) and synaptic feedback ($\mathbf{B}_l$) ensures learning efficiency comparable to backpropagation. (a) Alignment angle between forward and backward weights using various training methods (purple: $\mathbf{W}_l$ initialized with $\mathbf{B}_l^T$) and its corresponding validation accuracy (b).}
	\label{figa10}
\end{figure}

Our results indicate that pretraining with random noise effectively pre-aligns the forward weights with the backward weights, enhancing learning efficiency during subsequent data training. To support our claims, we conducted additional experiments using the CIFAR-10 dataset, which further validated our model. Specifically, we initialized the forward weights $\mathbf{W}_l$ to match the transpose of backward synaptic feedback $\mathbf{B}_l$ and proceeded with training (Figure \ref{figa10}. This approach yielded gradients and alignment similar to those observed in backpropagation during training. Consequently, we confirmed that this setup demonstrates learning efficiency comparable to that of backpropagation.

During the learning process, particularly in the initial stages, weight alignment may slightly degrade but largely remains intact. This baseline example illustrates that achieving initially aligned forward and backward weights allows for learning that is comparable to backpropagation, even without the enforced synchronization of backward weights during training. This serves as a specific instance highlighting the validity of random noise training in aligning forward and backward weights in a biologically plausible manner prior to data-driven learning.

\newpage
\subsection{Computational benefits of random noise training considering total training epochs}

\begin{figure}[H]
	\centering
	\includegraphics[width=0.75\textwidth]{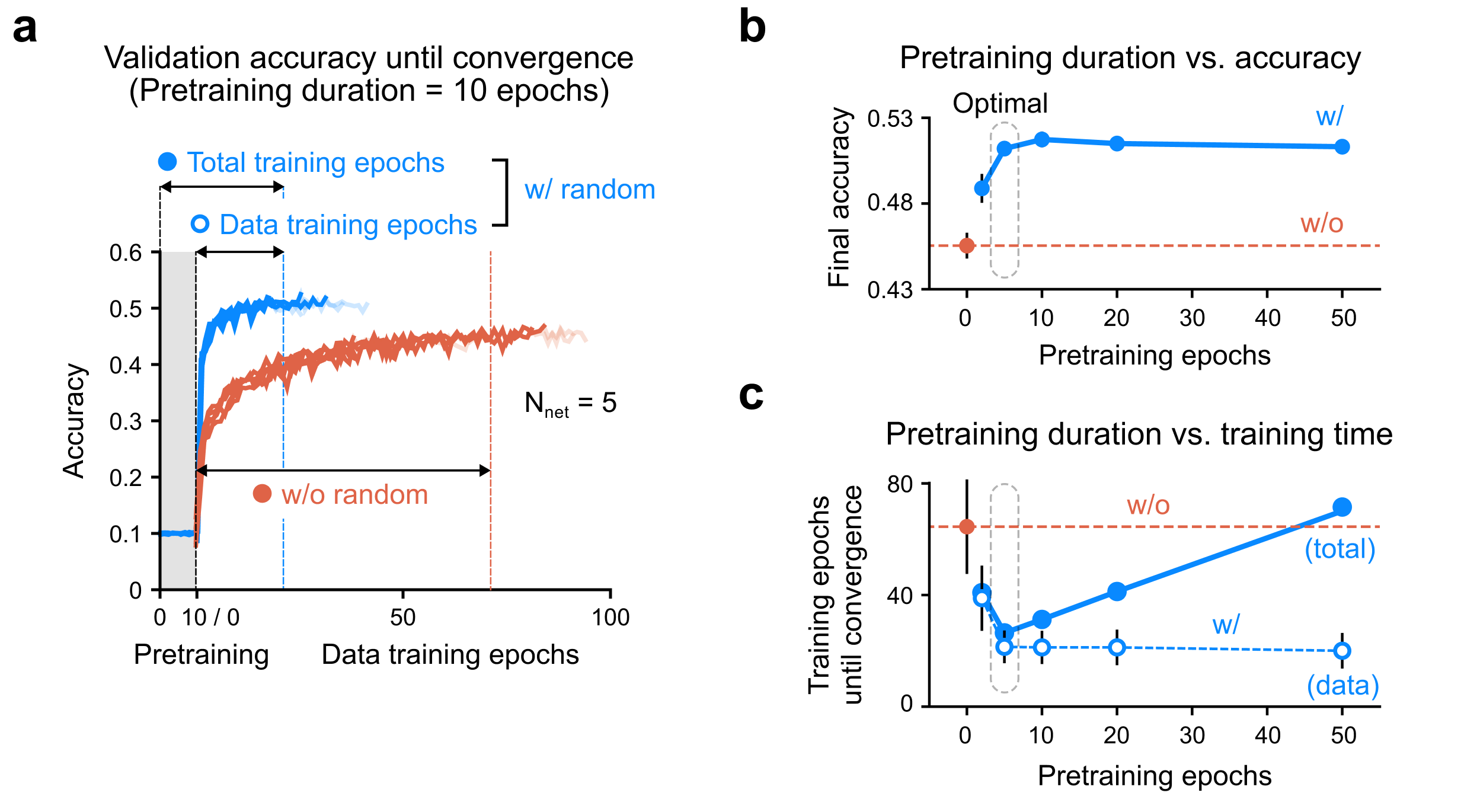}
	\caption{Comparison of training time until convergence including pretraining.
 (a) Learning curves of a four-layer MLP pretrained for 10 epochs and subsequently trained on CIFAR-10.
 (b-c) Repeated experiments with different pretraining durations.
 (b) Accuracy.
 (c) Training epochs until convergence.
 }
	\label{figa11}
\end{figure}

We confirmed that random noise pretraining accelerates convergence and can reduce computational resources, even when considering the total training duration (pretraining + data training) (Figure \ref{figa11}). By varying the duration of noise training, we demonstrated that, despite the additional time required for noise training, the total training time remained significantly shorter than that of training with data alone in most conditions. We conducted subsequent data training on networks pretrained with random noise for 2, 5, 10, 20, and 50 epochs, measuring the epochs required for training to converge (when validation accuracy no longer increased, with patience of 10 epochs). We maintained consistency in the number of samples used per epoch during both random noise training and subsequent data training, ensuring direct comparability of epoch times.

We found that longer periods of random noise training resulted in shorter subsequent data training times to achieve convergence. As reported previously, when comparing data training alone, the learning time was consistently much shorter in networks pretrained with noise. Notably, even when the time for noise training was included, the overall training duration for the noise-training algorithm remained substantially shorter than that for training with data alone in most conditions. Based on this analysis, we estimated an optimal duration for random noise pretraining that ensures the most efficient use of resources (Figure \ref{figa11}b, c, Optimal). We also found that at this optimal duration, the improvement in accuracy remained significant. These additional experimental results demonstrate that, when considering the resources expended in random noise training, it represents a more efficient approach (reduced training time and overall computation) compared to training with data alone.

\newpage
\section{Experimental details and additional results for section 4.3}

\subsection{Network architecture and training details}

\begin{table}[H]
    \caption{Parameters and settings used in the experiment.}
    \label{tablea6}
    \centering
    \begin{tabular}{lc}
        \toprule
        Name                      & Setting                 \\
        \midrule
        Dimensions                & {[}784, 100, 100, 10{]}     \\
        Activation function       & ReLU                    \\
        Number of random noise inputs        & $5 \times 10^5$         \\
        Batch size                & 64                      \\
        Learning rate             & 0.0001                  \\
        Optimizer                 & Adam                    \\
        \bottomrule
    \end{tabular}
\end{table}

In Section 4.3, we utilized a three-layer feedforward neural network for classification, comprising 100 neurons in the hidden layer and employing the ReLU as the activation function. Detailed hyperparameters of the network architecture and training can be found in Table \ref{tablea6}.

\subsection{Random noise training reduces the effective dimensionality of weights}

\begin{figure}[H]
    \centering
    \vspace{-0.3cm}
    \includegraphics[width=0.75\textwidth]{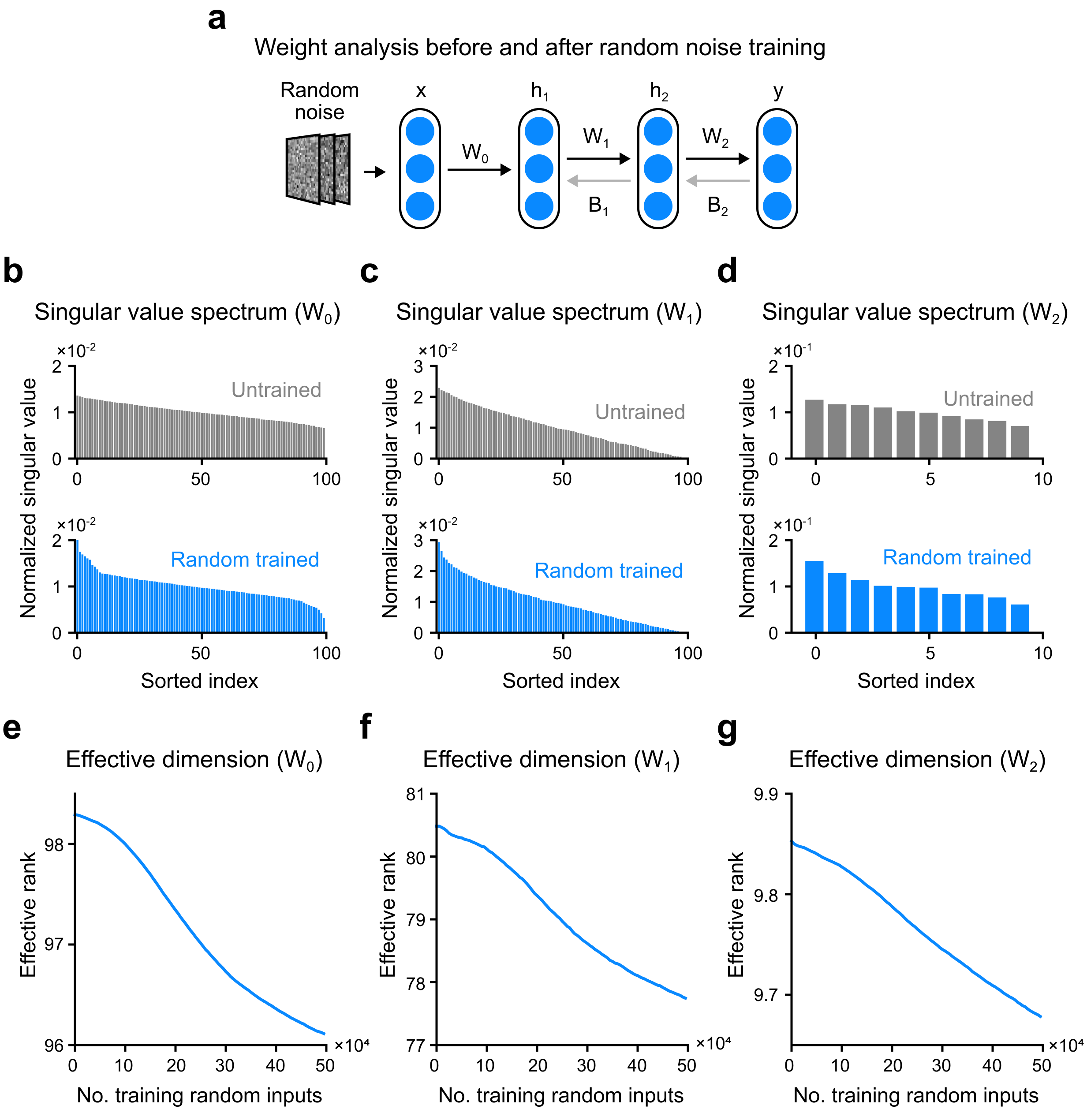}
    \vspace{-0.3cm}
    \caption{Effective dimensionality of weights in various layers. (a) Architecture of the network used in the experiment, in which the weights of the three layers are analyzed. (b-d) Singular value spectrum of weights in each layer of the untrained and randomly trained networks. (e-g) Effective dimensionality of weights in each layer during random noise training.}
    \label{figa12}
\end{figure}

We demonstrated that training with random noise effectively reduces the dimensionality of the weights, resulting in the learning of simpler solutions with lower rank. In Figure 4, we only illustrated the decrease in dimensionality in the first layer. Figure \ref{figa12} shows that the same trend is also observed in the remaining layers.

\newpage
\subsection{Enhanced generalization with various training sizes}

\begin{table}[H]
    \caption{Parameters and settings used in the experiment.}
    \label{tablea7}
    \centering
    \begin{tabular}{lc}
        \toprule
        Name                      & Setting                    \\
        \midrule
        Dimensions                & {[}784, 100, 100, 10{]}    \\
        Activation function       & ReLU                       \\
        Number of training data   & [100, 200, 400, 800, 1600] \\
        Number of test data       & $1 \times 10^3$            \\
        Epochs                    & 500                        \\
        Batch size                & 64                         \\
        Learning rate             & 0.0001                     \\
        Optimizer                 & Adam                       \\
        \bottomrule
    \end{tabular}
\end{table}

\begin{figure}[H]
    \centering
    \includegraphics[width=\textwidth]{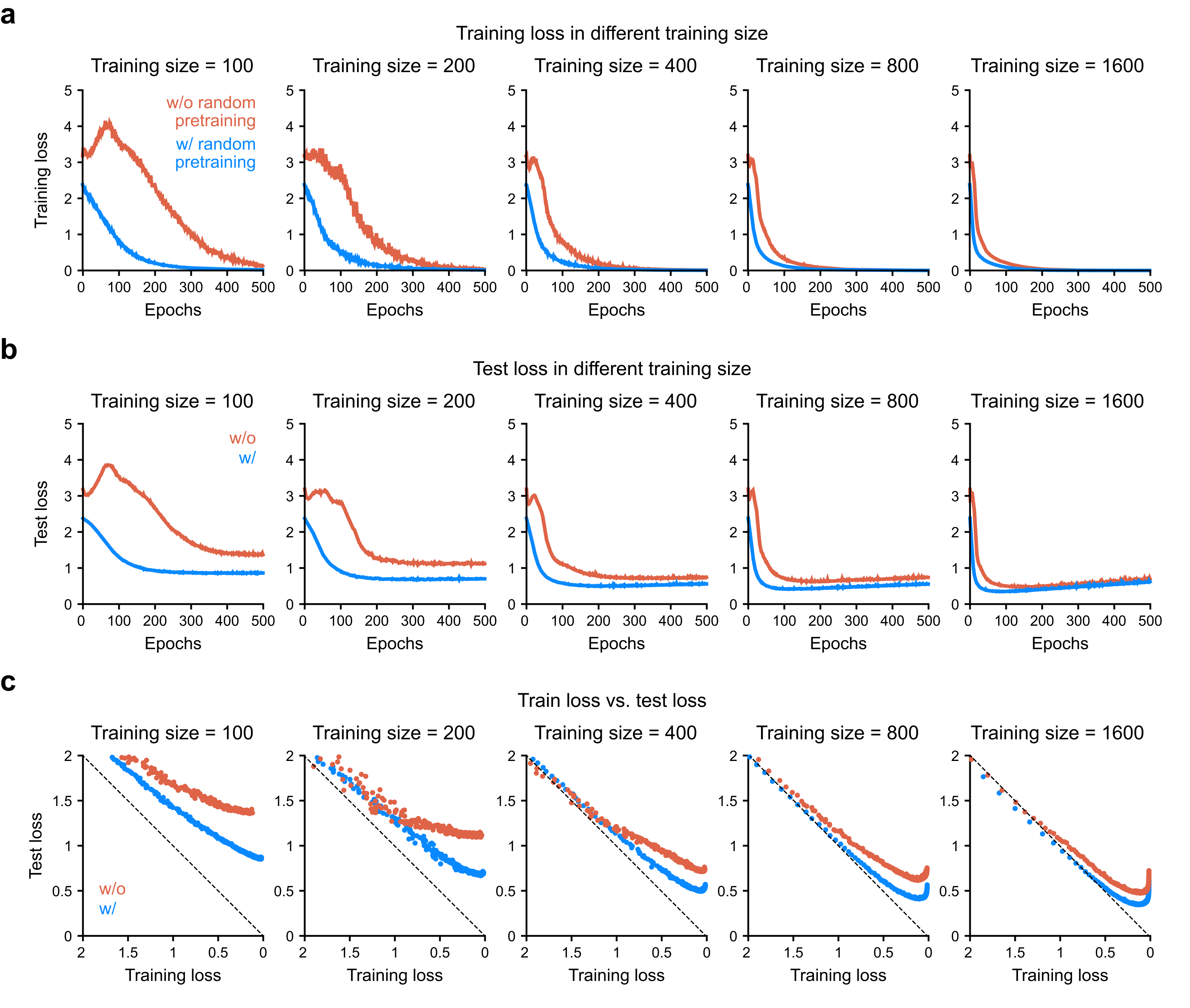}
    \caption{Training process and generalization error for different training set sizes. (a) Training loss. (b) Test loss. (c) Comparison of training loss versus test loss. Each column represents a network trained with a different training set size, with orange lines indicating networks trained solely on data and blue lines representing randomly pretrained networks.}
    \label{figa13}
\end{figure}

\newpage

\begin{figure}[H]
    \centering
    \includegraphics[width=\textwidth]{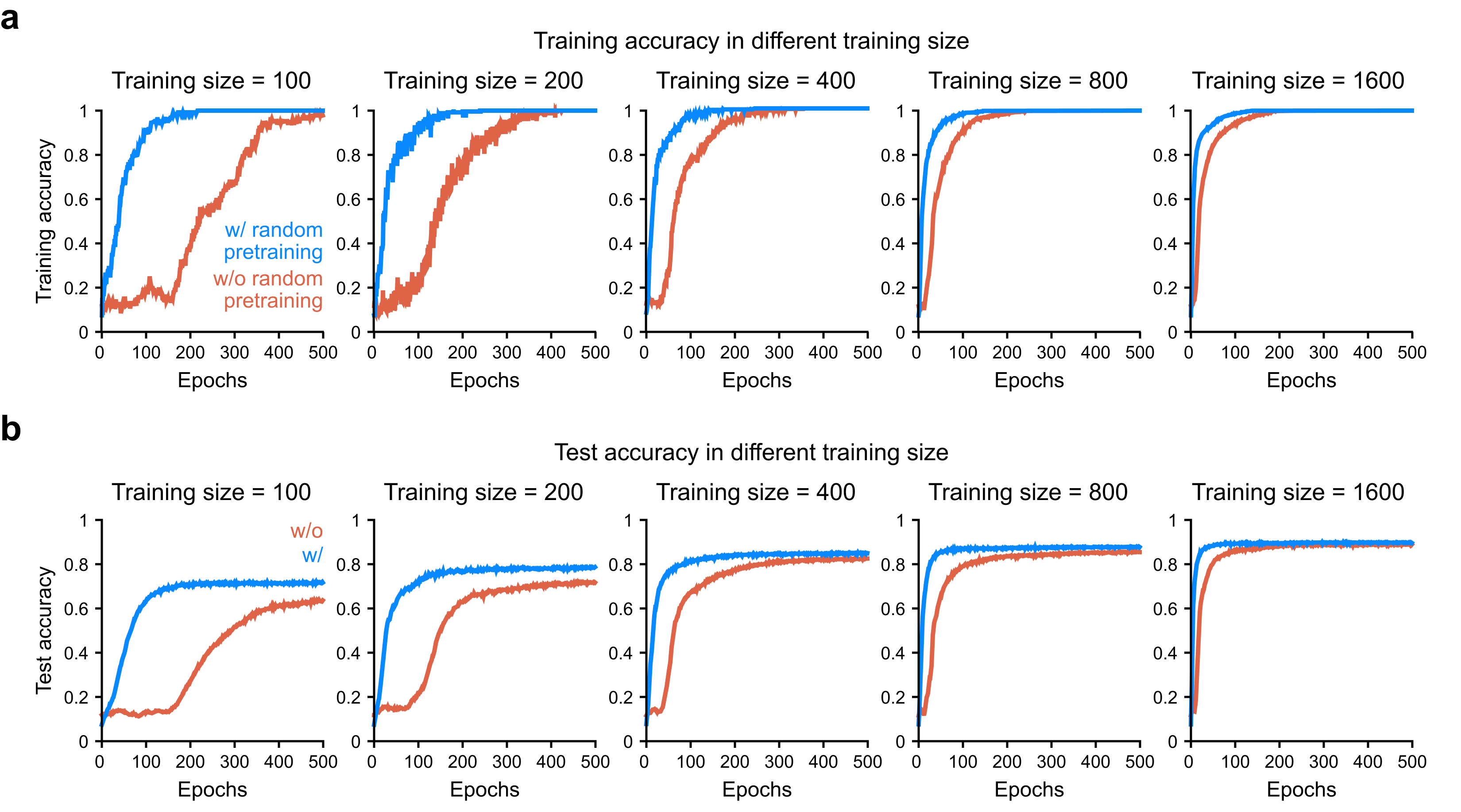}
    \caption{Training and test accuracy for different training set sizes. (a) Training accuracy. (b) Test accuracy. Each column represents a network trained with a different training set size, with orange lines indicating networks trained solely on data and blue lines representing randomly pretrained networks.}
    \label{figa14}
\end{figure}

We systematically analyzed the effect of random noise pretraining on generalization across various training sizes. Specifically, we trained three-layer neural networks with different training data sizes, including 100, 200, 400, 800, and 1600 samples. Detailed hyperparameters of the network architecture and training can be found in Table \ref{tablea7}. Across the different training sizes, we measured the generalization error (Figure \ref{figa13}) and accuracy (Figure \ref{figa14}). We observed that the generalization error was significantly reduced in networks pretrained with random noise, regardless of the training size. The summarized results are displayed in Figure 4f.

\newpage
\subsection{Enhanced generalization with various network depths}

\begin{table}[H]
    \caption{Parameters and settings used in the experiment.}
    \label{tablea8}
    \centering
    \begin{tabular}{lc}
        \toprule
        Name                         & Setting                 \\
        \midrule
        Input, hidden layer, output dimensions  & 784, 100, 10 \\
        Number of hidden layers      & [3, 4, 5, 6, 7]         \\
        Activation function          & ReLU                    \\
        Number of training data   & 1600                       \\
        Number of test data       & $1 \times 10^3$            \\
        Epochs                    & 500                        \\
        Batch size                & 64                         \\
        Learning rate             & 0.0001                     \\
        Optimizer                 & Adam                       \\
        \bottomrule
    \end{tabular}
\end{table}

\begin{figure}[H]
    \centering
    \includegraphics[width=\textwidth]{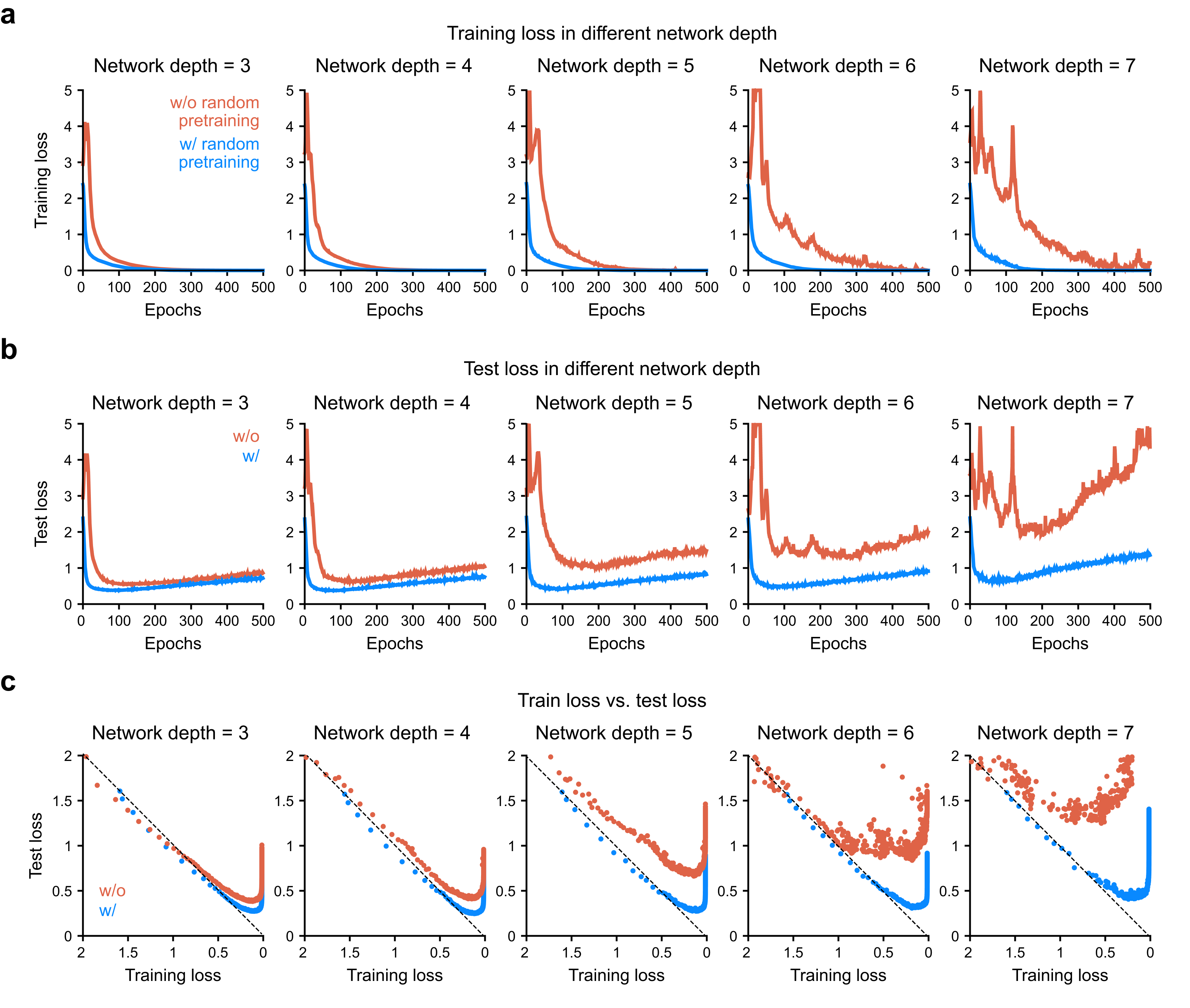}
    \caption{Training process and generalization error for different network depths. (a) Training loss. (b) Test loss. (c) Comparison of training loss versus test loss. Each column represents a network trained with a different training set size, with orange lines indicating networks trained solely on data and blue lines representing randomly pretrained networks.}
    \label{figa15}
\end{figure}

\newpage

\begin{figure}[H]
    \centering
    \includegraphics[width=\textwidth]{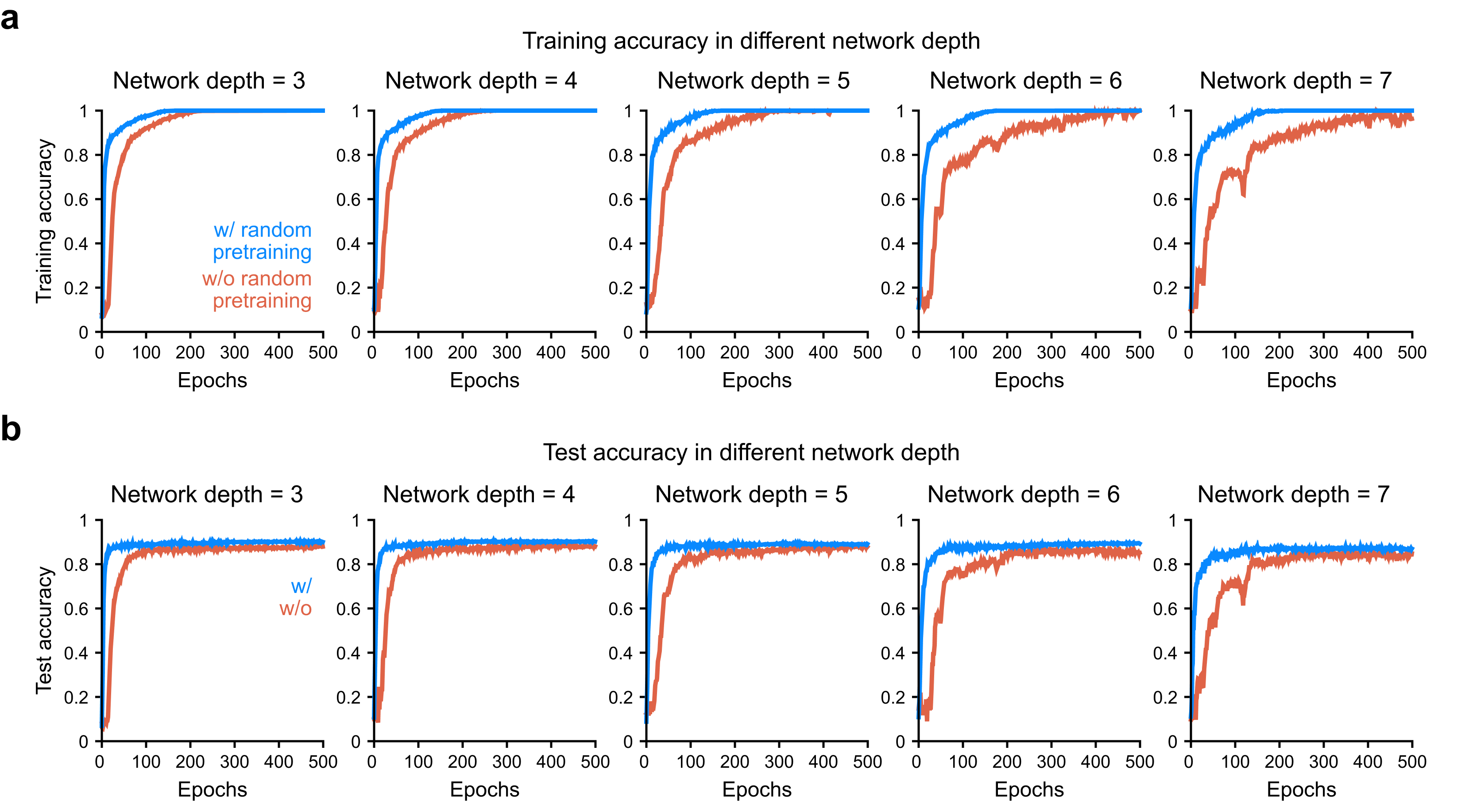}
    \caption{Training and test accuracy with different network depths. (a) Training accuracy. (b) Test accuracy. Each column represents a network trained with a different training set size, with orange lines indicating networks trained solely on data and blue lines representing randomly pretrained networks.}
    \label{figa16}
\end{figure}

Next, we analyzed the effect of random noise pretraining on generalization across various network depths. In this experiment, we fixed the training data size and varied the number of hidden layers. We tested three-, four-, five-, six-, and seven-layer neural networks with a hidden layer size of 100. Detailed hyperparameters of the network architecture and training can be found in Table \ref{tablea8}. Across the different network depths, we measured the generalization error (Figure \ref{figa15}) and accuracy (Figure \ref{figa16}). We observed that the generalization error was significantly reduced in networks pretrained with random noise, regardless of the network depth. Additionally, we assessed the effective rank of the learned features in this experiment and confirmed that random noise pretraining significantly reduces the dimensionality of the solution (Figure 4g, h). This may provide an explanatory scenario for how random noise pretraining enhances generalization across different network depths.

\newpage
\subsection{Generalization test for out-of-distribution}

\begin{figure}[H]
    \centering
    \includegraphics[width=\textwidth]{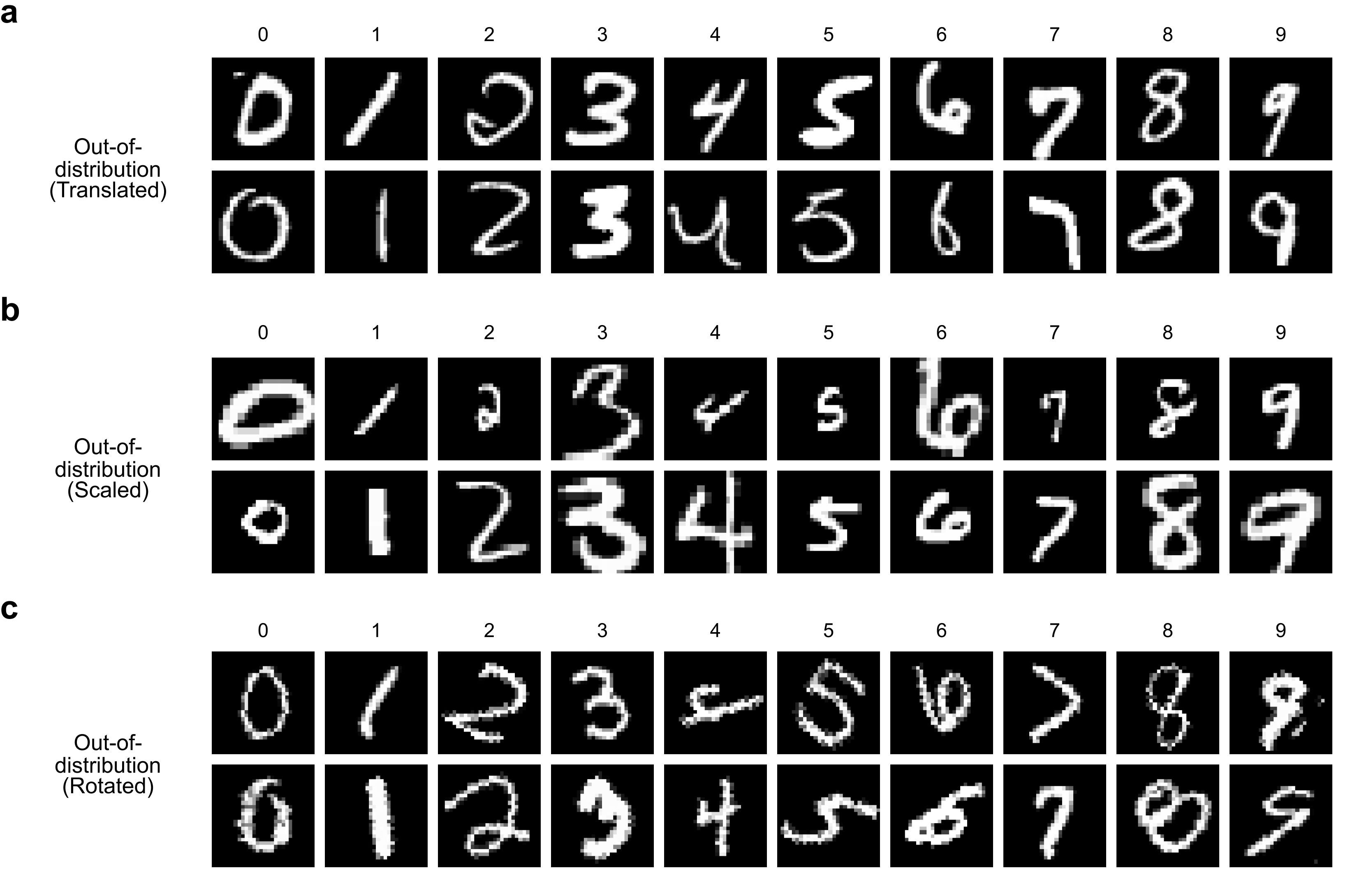}
    \caption{Datasets used in the out-of-distribution generalization test with the transformed dataset.
    (a) MNIST dataset used in the training in-distribution case.
    (b-d) Transformed MNIST dataset utilized in the generalization test for the out-of-distribution case.
    (b) Translated MNIST dataset, where each image is randomly translated in the range of $[-5\%, 5\%]$ of the input image size on the x- and y-axis.
    (c) Scaled MNIST dataset, where each image is randomly scaled in the range of $[0.8, 1.2]$.
    (d) Rotated MNIST dataset, where each image is randomly rotated in the range of $[-25, 25]$ degrees.
    }
    \label{figa17}
\end{figure}

\begin{figure}[H]
    \centering
    \includegraphics[width=\textwidth]{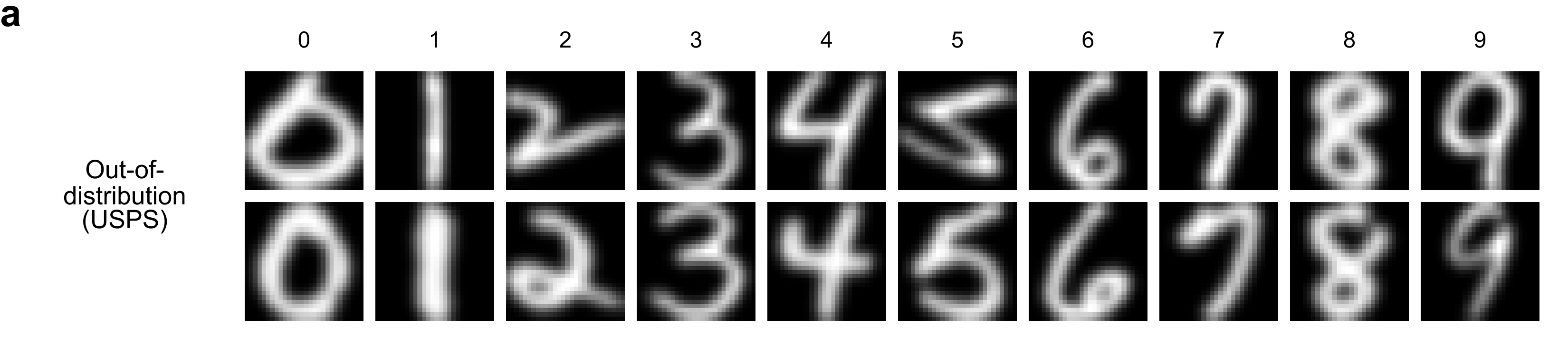}
    \caption{Dataset used in the out-of-distribution generalization test with the benchmark dataset.
    (a) MNIST dataset \cite{deng2012} used to train the networks in the out-of-distribution case.
    (b) USPS dataset \cite{hull1994} resized from $(16 \times 16)$ to $(28 \times 28)$, utilized in the generalization test for the out-of-distribution case.}
    \label{figa18}
\end{figure}

We investigated whether networks trained with random noise can generalize to out-of-distribution data. To assess the model’s performance on transformed MNIST, we created a custom dataset by applying translations, scaling, and rotations. Example data used in the experiment can be seen in Figure \ref{figa17}. The results are presented in Figure 5b. Additionally, we evaluated the model’s performance on a benchmark out-of-distribution generalization dataset, USPS (Figure \ref{figa18}). The corresponding results are shown in Figure 5c.

\newpage
\section{Experimental details and additional results for section 4.4}

\subsection{Network architecture and training details}

\begin{table}[H]
    \caption{Parameters and settings used in random noise pretraining.}
    \label{tablea9}
    \centering
    \begin{tabular}{lc}
        \toprule
        Name                      & Setting                 \\
        \midrule
        Dimensions                & {[}784, 100, 100, 10{]} \\
        Activation function       & ReLU                    \\
        Number of random noise inputs        & $5 \times 10^5$         \\
        Batch size                & 64                      \\
        Learning rate             & 0.0001                  \\
        Optimizer                 & Adam                    \\
        \bottomrule
    \end{tabular}
\end{table}

In Section 4.4, we employed a three-layer feedforward neural network for classification, comprising 100 neurons in the hidden layer and utilizing the ReLU as the activation function. Detailed hyperparameters of the network architecture and random noise training can be found in Table \ref{tablea9}.

\begin{table}[H]
    \caption{Parameters and settings used in meta-loss measurement.}
    \label{tablea10}
    \centering
    \begin{tabular}{lc}
        \toprule
        Name                      & Setting                 \\
        \midrule
        K                         & 10                      \\
        Inner steps               & 10                      \\
        Inner learning rate             & 0.001                  \\
        Inner optimizer                 & Adam                   \\
        Task distribution (Dataset)     & MNIST, F-MNIST, K-MNIST \\
        \bottomrule
    \end{tabular}
\end{table}

During training with random noise, we measured the meta-loss, which evaluates the network’s ability to quickly adapt to various tasks. To compute the meta-loss, we utilized the definition from model-agnostic meta-learning (MAML) \cite{finn2017}, a widely used meta-learning algorithm. We constructed a task distribution consisting of three datasets: MNIST, Fashion-MNIST, and Kuzushiji-MNIST. We sampled batches of tasks from these datasets and trained the copied networks on the sampled tasks for a few gradient steps. Subsequently, we summed the loss from the adapted networks and used this sum as the meta-loss. Detailed hyperparameters for the meta-loss measurement can be found in Table \ref{tablea10}. Unlike conventional meta-learning approaches such as MAML — where the meta-loss is employed to optimize the network’s initial parameters for rapid adaptation — we simply measured the meta-loss during random noise training without using it to update the network.

\begin{table}[H]
    \caption{Parameters and settings used in task adaptation.}
    \label{tablea11}
    \centering
    \begin{tabular}{lc}
        \toprule
        Name                      & Setting                 \\
        \midrule
        Number of training data   & $5 \times 10^3$         \\
        Number of test data       & $5 \times 10^3$         \\
        Epochs                    & 100                     \\
        Batch size                & 64                      \\
        Learning rate             & 0.0001                  \\
        Optimizer                 & Adam                   \\
        Task distribution (Dataset)     & MNIST, F-MNIST, K-MNIST \\
        \bottomrule
    \end{tabular}
\end{table}

Next, we compared untrained networks with networks pretrained using random noise in terms of their adaptation to each task. We copied the networks and adapted them to three tasks: MNIST, Fashion-MNIST, and Kuzushiji-MNIST. Detailed hyperparameters used in task adaptation can be found in Table \ref{tablea11}.

\newpage
\subsection{Measurement of meta-loss}

\begin{figure}[H]
    \centering
    \includegraphics[width=\textwidth]{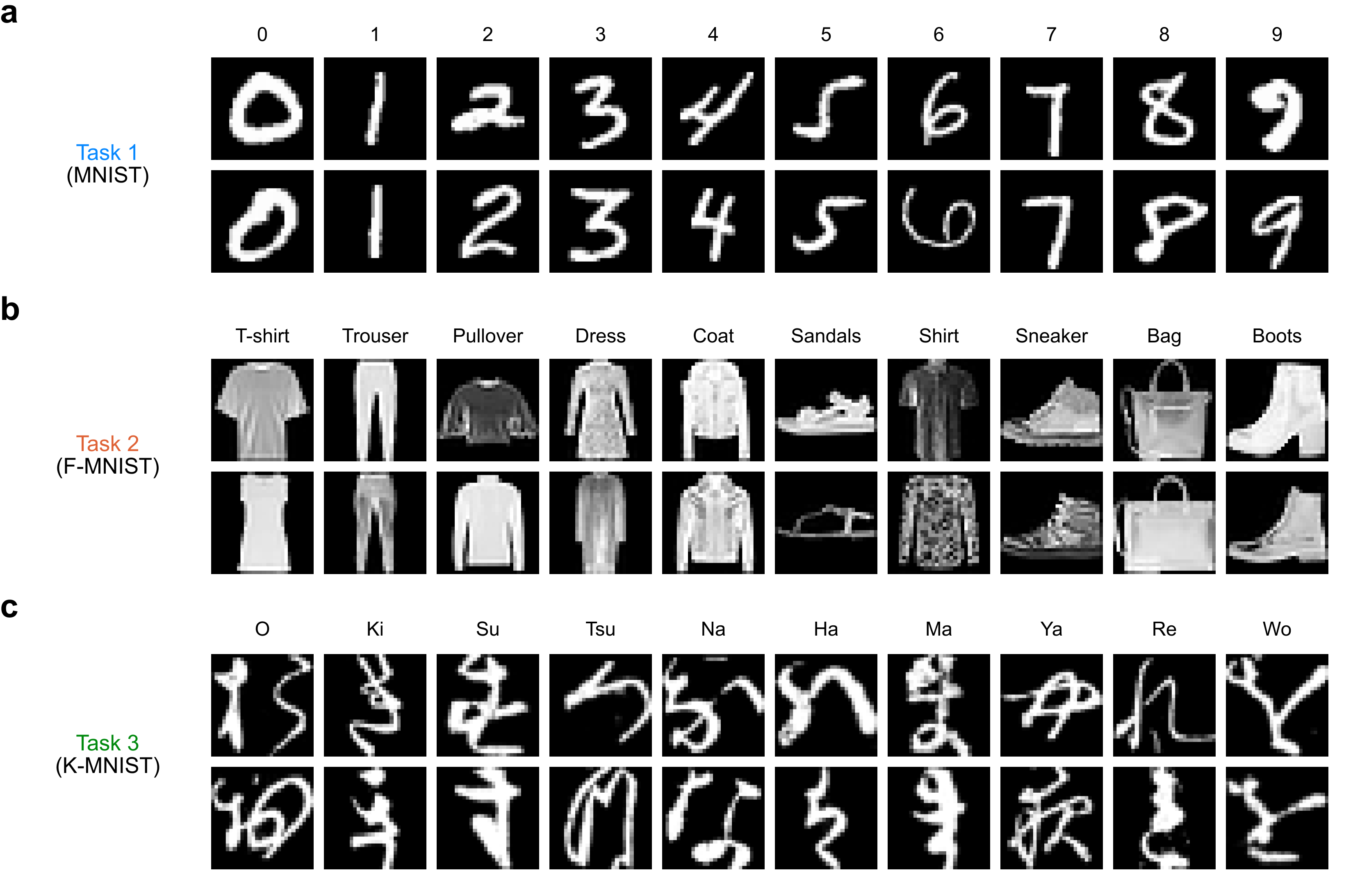}
    \caption{Tasks used to measure the meta-loss.
    (a) Task 1: MNIST \cite{deng2012} classification task.
    (b) Task 2: Fashion-MNIST \cite{xiao2017} classification task.
    (c) Task 3: Kuzushiji-MNIST \cite{clanuwat2018} classification task.}
    \label{figa19}
\end{figure}

\begin{figure}[H]
    \centering
    \includegraphics[width=0.5\textwidth]{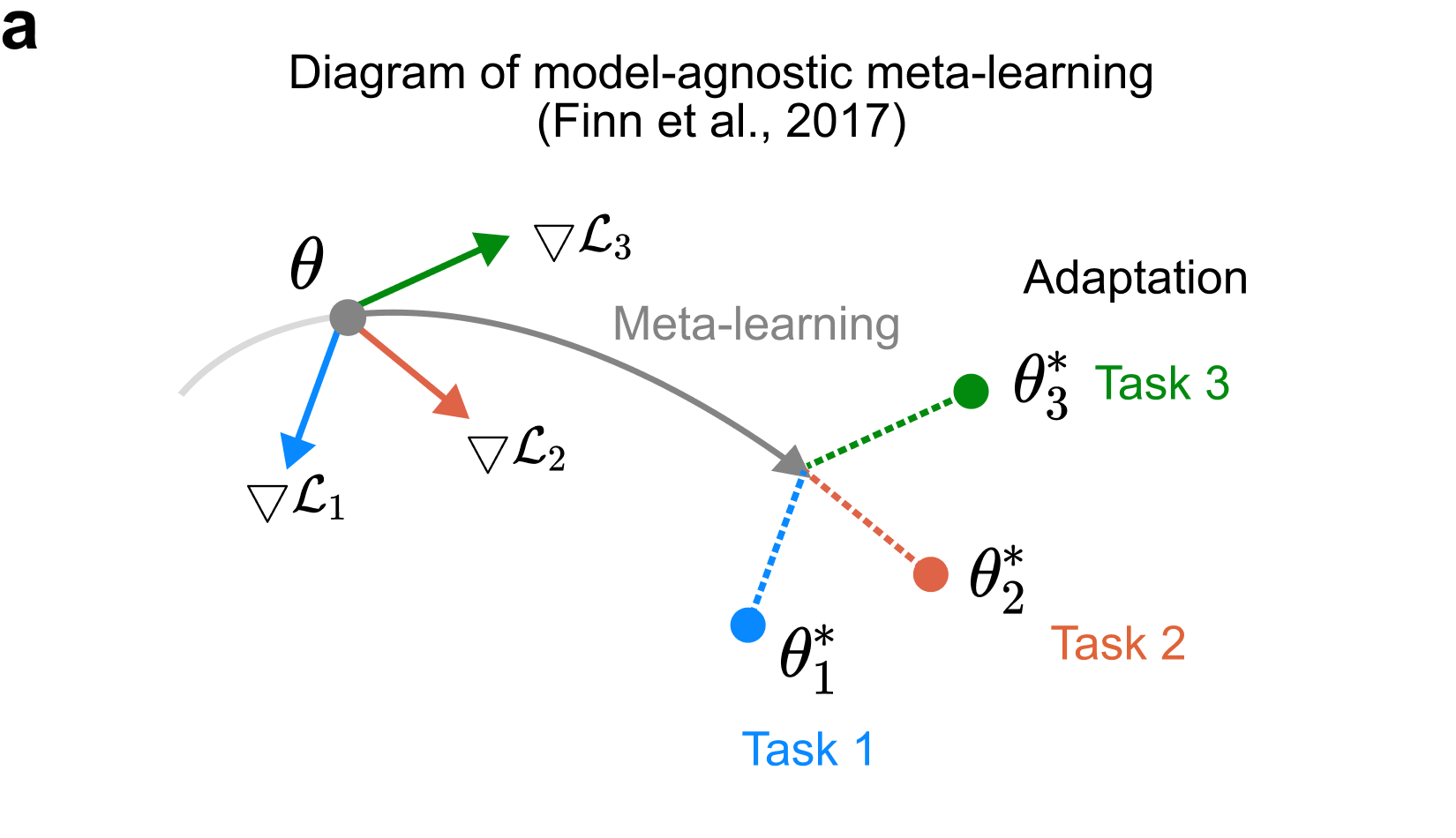}
    \caption{Diagram of meta-learning. Each colored arrow indicates the gradient update direction corresponding to a specific task. The gray arrow represents the overall learning direction of the meta-learning process. Each colored point denotes adaptation to a specific task from the meta-learned network.}
    \label{figa20}
\end{figure}

We considered three tasks: MNIST classification, Fashion-MNIST classification, and Kuzushiji-MNIST classification (Figure \ref{figa19}) to measure the meta-loss and assess the fast adaptation of networks pretrained with random noise. The diagram of the meta-learning process (Figure \ref{figa20}) illustrates that optimizing the network to minimize the meta-loss enables quicker adaptation to task distributions. Notably, our results (Figure 6a) demonstrate that random noise pretraining efficiently reduces the meta-loss without the need for exposure to task-specific data. Additionally, the trajectory of weights in latent space during adaptation to various tasks (Figure 6b) follows a pattern similar to that depicted in the diagram. This highlights the role of random noise pretraining as a form of meta-learning that facilitates rapid adaptation.

\newpage
\section{Our contributions}

Our goal is to enhance current strategies in deep learning by drawing insights from brain function. There exists a significant accuracy gap between backpropagation and biologically plausible learning strategies that do not involve weight transport. While backpropagation is effective, it is computationally intensive, requiring dynamic memory access for weight transport (i.e., accessing forward weights in memory to compute backward updates). Pretraining with random noise and aligning forward-backward weights achieves similar outcomes without relying on weight transport. Our interest is not solely in attaining performance comparable to backpropagation, but rather in achieving this without weight transport. Feedback alignment (and random noise pretraining) does not require weight transport and relies exclusively on local information. In the context of energy-efficient neuromorphic chip engineering, feedback alignment is sometimes employed for learning, albeit with some performance trade-offs. Our results demonstrate that such sacrifices are minimal in our model. Overall, our findings provide insights into how to reach levels of learning efficiency similar to backpropagation while circumventing the need for weight transport.


\section{Experimental details}

\textbf{Data availability.} The datasets used in this study are publicly available: \url{http://yann.lecun.com/exdb/mnist/} (MNIST \cite{deng2012}), \url{https://github.com/zalandoresearch/fashion-mnist} (Fashion-MNIST \cite{xiao2017}), \url{https://github.com/rois-codh/kmnist} (Kuzushiji-MNIST \cite{clanuwat2018}), \url{https://www.csie.ntu.edu.tw/~cjlin/libsvmtools/datasets/multiclass.html#usps} (USPS \cite{hull1994}), \url{https://www.cs.toronto.edu/~kriz/cifar.html} (CIFAR-10 and CIFAR-100 \cite{krizhevsky2009}), \url{https://cs.stanford.edu/~acoates/stl10/} (STL-10 \cite{coates2011}).

\textbf{Software used.} Python 3.11 (Python software foundation) with PyTorch 2.1 and NumPy 1.26.0 was used to perform the simulation and the analysis. SciPy 1.11.4 was used to perform the statistical test and analysis. The code used in this work is available at \url{https://github.com/cogilab/Random}.

\textbf{Computing resources.} All simulations were performed on a computer with an Intel Core i7-11700K CPU and an NVIDIA GeForce GTX 1080 GPU. The simulation code was parallelized using PyTorch’s built-in parallelization to utilize the GPU resources efficiently.
